\documentclass[10pt,twocolumn,letterpaper]{article}

\usepackage{cvpr}              

\usepackage{tikz}
\usepackage{dblfloatfix}
\usepackage{pgfplots}
\pgfplotsset{compat=1.16}
\usepgfplotslibrary{groupplots}
\usetikzlibrary{patterns}
\usepackage[T1]{fontenc}
\usepackage{tablefootnote}
\usepackage{pgfkeys}
	\def\addlegendimage{\csname pgfplots@addlegendimage\endcsname}
\usetikzlibrary{arrows}
\usepackage{tkz-kiviat}
\usepackage{xfp}
\usepackage{graphicx}
\usepackage{amsmath}
\usepackage{amssymb}
\usepackage{booktabs}
\usepackage{textcomp}
\usepackage{siunitx}

\definecolor{citecolor}{HTML}{0071bc}
\definecolor{color_ao}{gray}{0.5}
\definecolor{color_our}{rgb}{0.66,0.82,0.56}
\definecolor{color_pre}{rgb}{0.52,0.59,0.69}
\definecolor{Gray}{gray}{0.9}
\definecolor{LighterGray}{gray}{0.93}
\definecolor{LightGrayForTableRule}{gray}{0.92}
\definecolor{DarkGray}{gray}{0.5}
\definecolor{Black}{rgb}{0.0, 0.0, 0.0}
\definecolor{NiceBlue}{rgb}{0.11764705882352941, 0.5647058823529412, 1.0}
\definecolor{NiceGreen}{HTML}{3db0fc}
\definecolor{cvprblue}{rgb}{0.21,0.49,0.74}
\usepackage[pagebackref,breaklinks,colorlinks,allcolors=cvprblue]{hyperref}
\usepackage[accsupp]{axessibility} 
\usepackage{array}
 \usepackage{multirow}
 \usepackage[table,xcdraw]{xcolor}
 \usepackage{listings}
\usepackage{xcolor}
\usepackage[most]{tcolorbox}
\usepackage{minitoc}
\usepackage{subcaption}
\newcommand{\nameemoji}{\includegraphics[height=.8\baselineskip]{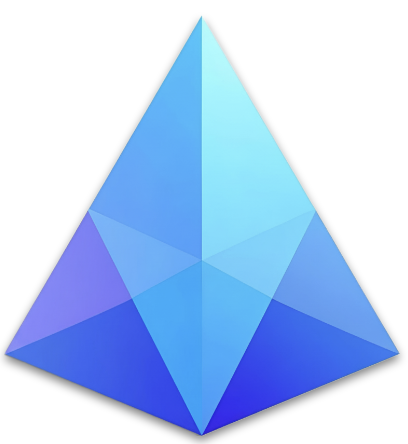} PRISM}

\def\paperID{12074} 
\def\confName{CVPR}
\def\confYear{2026}

\title{Through the \nameemoji{}: Principle-Aware, Interpretable, and Multi-Scale Evaluation of Visual Designs}

\author{Mona Gandhi$^1$*, K J Joseph$^2$, Srinivasan Parthasarathy$^1$, Sayan Nag$^2$\\$^1$Ohio State University, $^2$Adobe Research\\\texttt{\{gandhi.255, srini\}@osu.edu, \{josephkj, snag\}@adobe.com}
}

\begin{document}
\maketitle

\def\thefootnote{*}\footnotetext{Work done at Adobe Research}\def\thefootnote{\arabic{footnote}}

\begin{abstract}
Effective visual communication stems from the harmony of multiple design principles, such as readability, contrast, alignment, overlap, and coherence, which collectively govern clarity and intent of the communicator.
While human designers reason holistically over these principles, machine agents typically condense them into a single heuristic score, offering limited interpretability and diagnostic precision.
To address this gap, we introduce \textbf{\nameemoji{}}  (\textbf{PR}inciple-aware, \textbf{I}nterpretable, and \textbf{S}tructure-guided Design \textbf{M}odifications), a benchmark that systematically perturbs professional layouts from the Crello dataset along measurable design principles. The benchmark comprises \textbf{100K} perturbed training samples and \textbf{10K} perturbed validation designs, each isolating a specific principle violation for controlled analysis of multimodal reasoning about design quality. We show that models like Qwen-2.5-VL and GPT-4o-mini are largely insensitive to targeted principle degradations, whereas GPT-4o exhibits global awareness without fine-grained disentanglement.
Building on these insights, we propose a multi-scale evaluation framework that integrates lightweight scorers for quantitative assessment, instruction-tuned vision-language models for localised feedback, and prompt-based methods for global reasoning.
Our framework provides interpretable explanations of design failures.
Using these localised insights, we show targeted refinements that improve layout quality.
Together, PRISM and our framework lay the foundation for interpretable design-literate multimodal reasoning systems.
\end{abstract}    
\begin{figure}[h!]
     \centering
     \includegraphics[width=\linewidth, trim={0cm 0cm 0cm 0cm},clip]{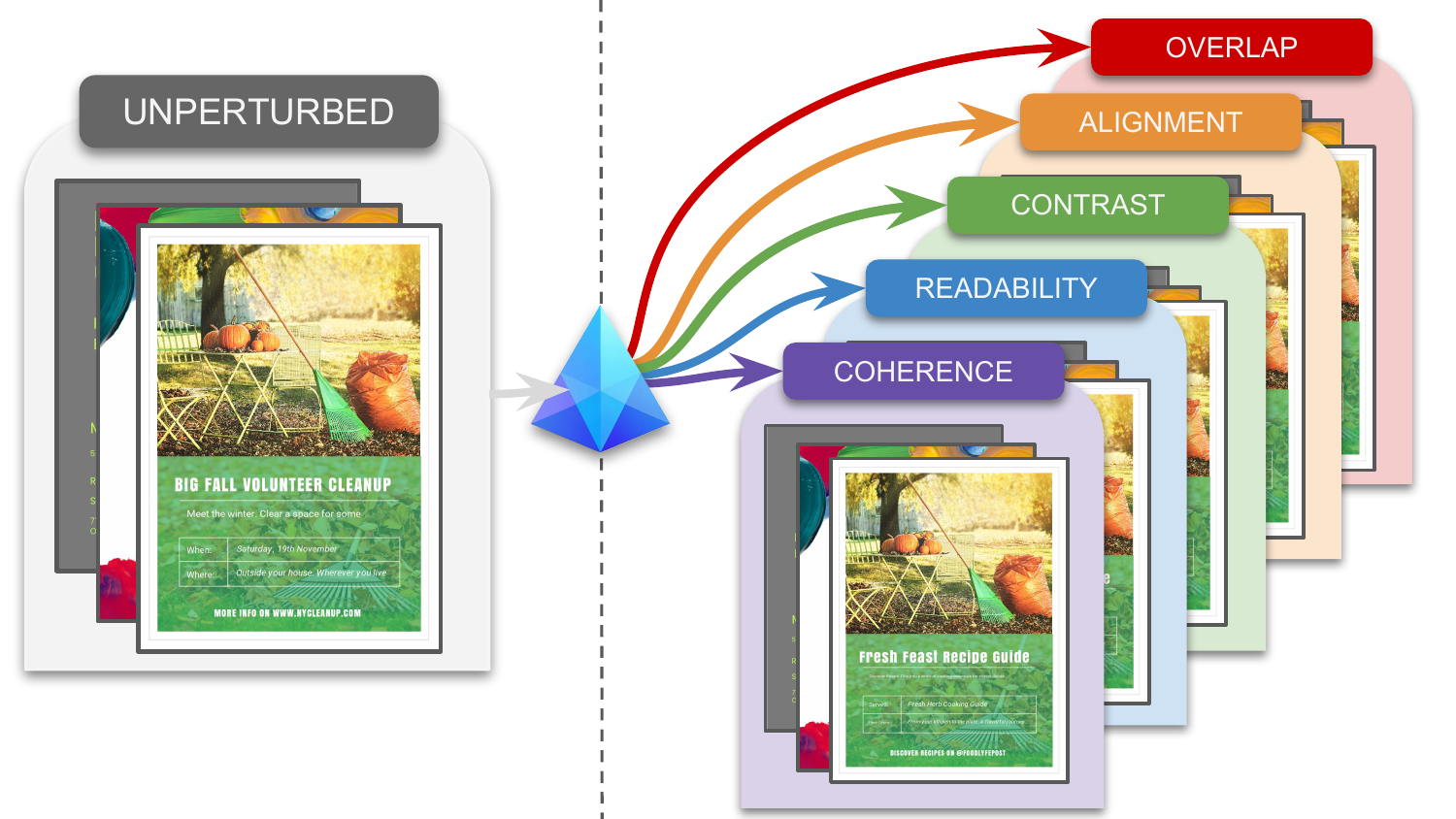}
     \caption{\textbf{We introduce \nameemoji{}}, a benchmark that operationalizes design evaluation through measurable principles of visual composition. Derived from professional layouts, PRISM systematically perturbs examples along five principles: \textit{coherence}, \textit{readability}, \textit{contrast}, \textit{alignment}, and \textit{overlap}, isolating targeted violations while preserving other attributes. These controlled perturbations enable interpretable and principle-specific evaluation of models’ sensitivity to localised design degradations and their holistic reasoning about visual quality.}
     \label{fig:pull_figure}
 \end{figure}
\section{Introduction}
\label{sec:intro}

Designing is inherently a process of reasoning about perception and communication. 
When evaluating a visual composition, human designers do not typically rely on a single notion of “good design”. 
They often weigh multiple interacting principles such as \textit{coherence}, \textit{readability}, \textit{contrast}, \textit{alignment}, \textit{balance}, and many more such design principles \cite{tacit}. 
A designer might praise a layout’s overall theme yet still adjust text spacing for legibility or modify color contrast for accessibility. 
This layered reasoning, which moves fluidly between global intent and local detail, enables designers to make targeted refinements preserving the composition’s purpose while enhancing clarity and aesthetics \cite{huang2023survey}.

In contrast, existing computational evaluation methods reduce this multifaceted process to a single scalar score, such as aesthetic preference or layout similarity~\cite{goyal2024design, otani2024ltsim, patnaik2025aesthetiqenhancinggraphiclayout}.  
While convenient, these metrics cannot reveal which aspects of a design are successful or unsuccessful, and they offer little guidance for actionable improvement \cite{sayan2024design, zou2025fragment} or provide end-to-end refiners \cite{goyal2024design}.  
Their aggregation of all principles into a single scalar is at odds with how human designers critique and refine, focusing on specific violations that can be iteratively addressed \cite{sayan2024design, vascar}.

Recent large language and multimodal models (LLMs and MLLMs) can produce free-form critiques that appear similar to human reasoning \cite{sayan2024design, duan2024uicrit, vascar}.  
However, their evaluations are stochastic, qualitative, and non-reproducible.  
Given the same poster, such models often generate inconsistent feedback, sometimes suggesting unnecessary global edits for well-designed layouts.  
Further, their responses lack grounding in explicit design principles and provide judgments without measurable justification. 
Thus, they behave more like subjective critics than consistent evaluators.

To bridge this gap, we introduce \textbf{\nameemoji{}} (\textbf{PR}inciple-aware, \textbf{I}nterpretable, and \textbf{S}tructure-guided Design \textbf{M}odifications), a curated perturbation dataset that systematically degrades professionally crafted designs from the Crello corpus \cite{canvasvae} along controlled dimensions of \textit{coherence}, \textit{readability}, \textit{contrast}, \textit{alignment}, and \textit{overlap} (ref Fig.~\ref{fig:pull_figure}).
Each perturbation isolates a targeted violation, for example, text color blending into the background for readability, or a mismatch between the headline and imagery for coherence, thus providing principle-specific supervision for multimodal evaluation.
These controlled degradations reflect how humans perceive drops in design quality: localised changes in text or contrast disrupt readability, while global inconsistencies undermine coherence and intent. Analysis on PRISM reveals a gap in current models: architectures like Qwen-2.5-VL \cite{qwen2_5} and GPT-4o-mini \cite{openai2024gpt4omini} often fail to register these controlled degradations, while GPT-4o \cite{gpt4o} shows broad semantic sensitivity but lacks consistent separation between fine-grained principles.

Building on PRISM, we propose a structured evaluation framework with principle-specific \textbf{scorers} and \textbf{localisers}.  
PRISM-scorer is a lightweight classifier built on a SigLIP-v2 \cite{siglipv2} backbone fine-tuned on design data for each principle, trained to produce quantitative scores.  
For local principles such as readability, we pair these scorers with an instruction-tuned Qwen-2.5-VL model that not only detects issues but also explains where they occur, similar to how a designer highlights specific regions needing adjustment.  
For higher-level properties such as coherence, which involve semantic and thematic consistency, we use prompting-based evaluation with MLLMs. Finally, we illustrate how the scorer and localiser enable interpretable, iterative refinement using a lightweight \textbf{beam search-based editing} \cite{shi2024thorough}. The editor proposes minimal adjustments that improve targeted principle scores while preserving the overall layout, offering controllable feedback beyond generative MLLMs \cite{bagel, gpt4o} that treat designs as static images.

\begin{figure}[h!]
     \centering
     \includegraphics[width=\linewidth, trim={0cm 0cm 0cm 0cm},clip]{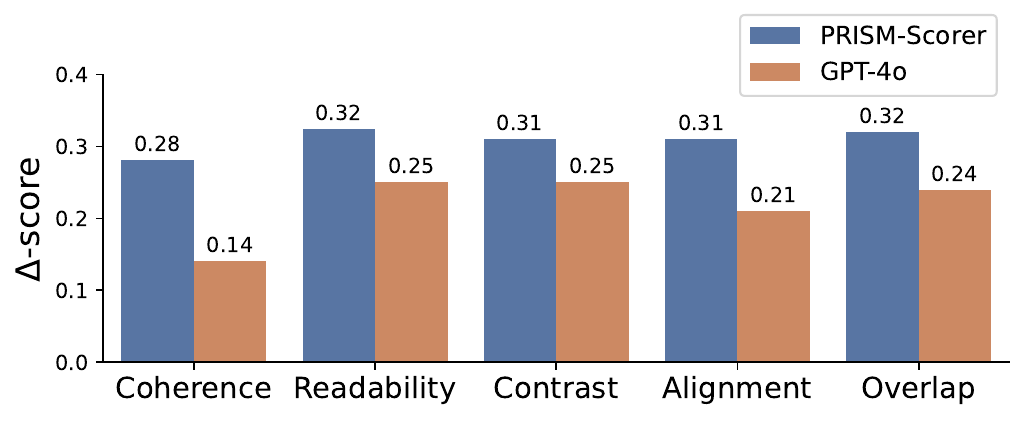}
     \caption{\textbf{PRISM-Scorer Performance.} $\Delta$-score bar plot comparing the sensitivity of different models to PRISM perturbations. Each axis shows the change in principle-specific scores between the original and perturbed posters, as assigned by each scorer. Higher values indicate stronger discrimination of principle violations. Refer to Section~\ref{sec:model_sensitivity} for further details.}
     \label{fig:main_radar}
 \end{figure}
\begin{figure*}[t]
\centering
\renewcommand{\arraystretch}{1.05}
\setlength{\tabcolsep}{4pt}

\begin{tabular}{ll|llllll}
\multicolumn{1}{c}{\textbf{Unperturbed}} & & & \multicolumn{1}{c}{\textbf{Coherence}} & \multicolumn{1}{c}{\textbf{Readability}} & \multicolumn{1}{c}{\textbf{Contrast}} & 
\multicolumn{1}{c}{\textbf{Alignment}} & \multicolumn{1}{c}{\textbf{Overlap}} \\
\includegraphics[width=0.14\linewidth]{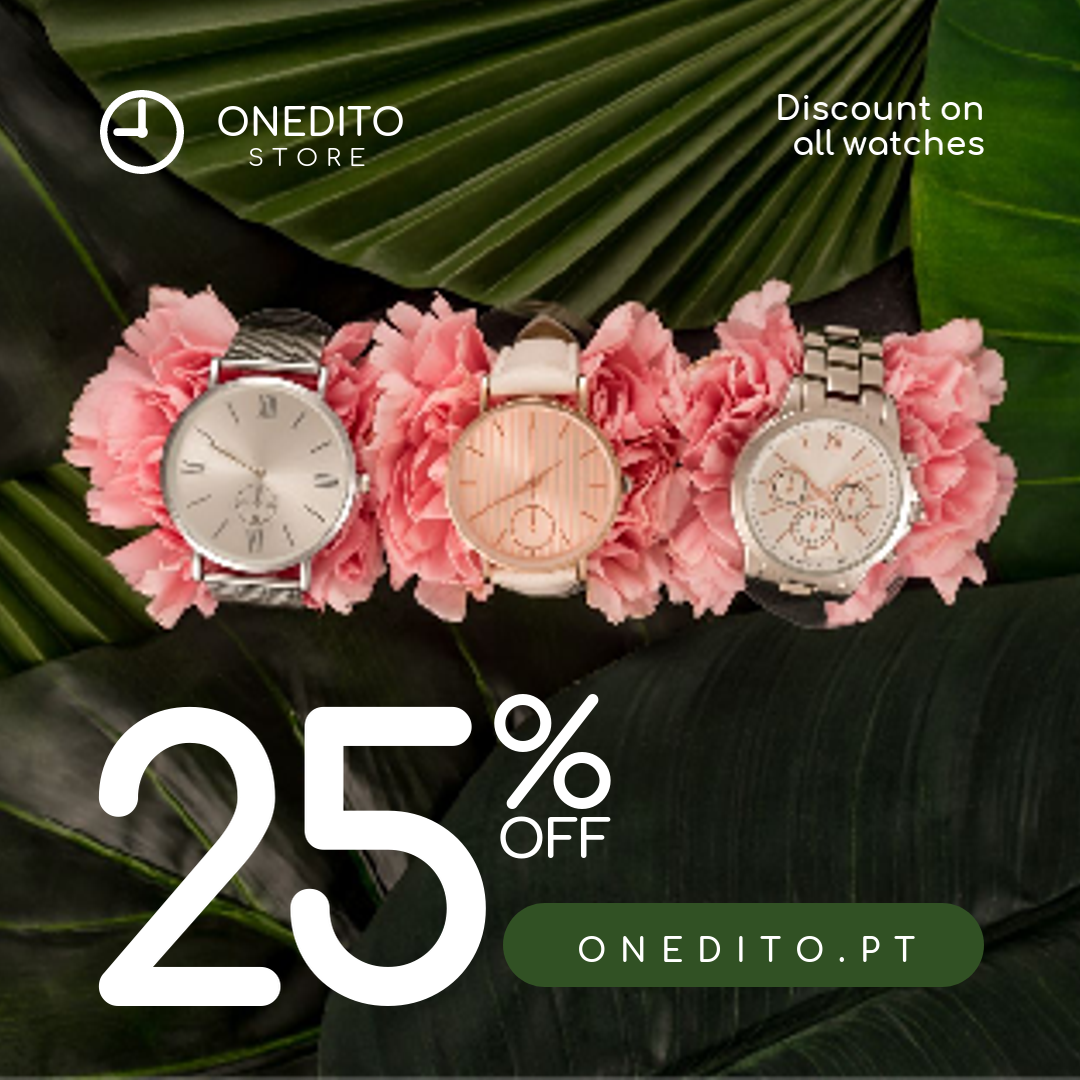} & & &
\includegraphics[width=0.14\linewidth]{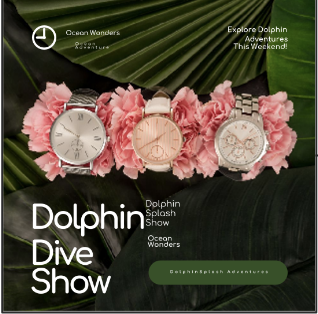} &
\includegraphics[width=0.14\linewidth]{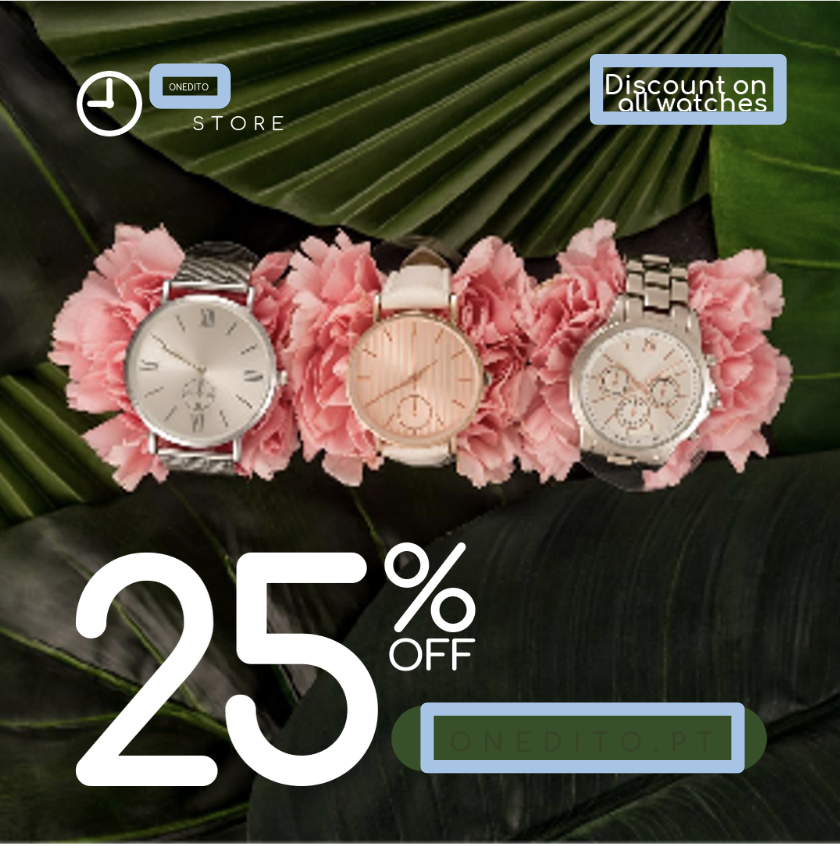} &
\includegraphics[width=0.14\linewidth]{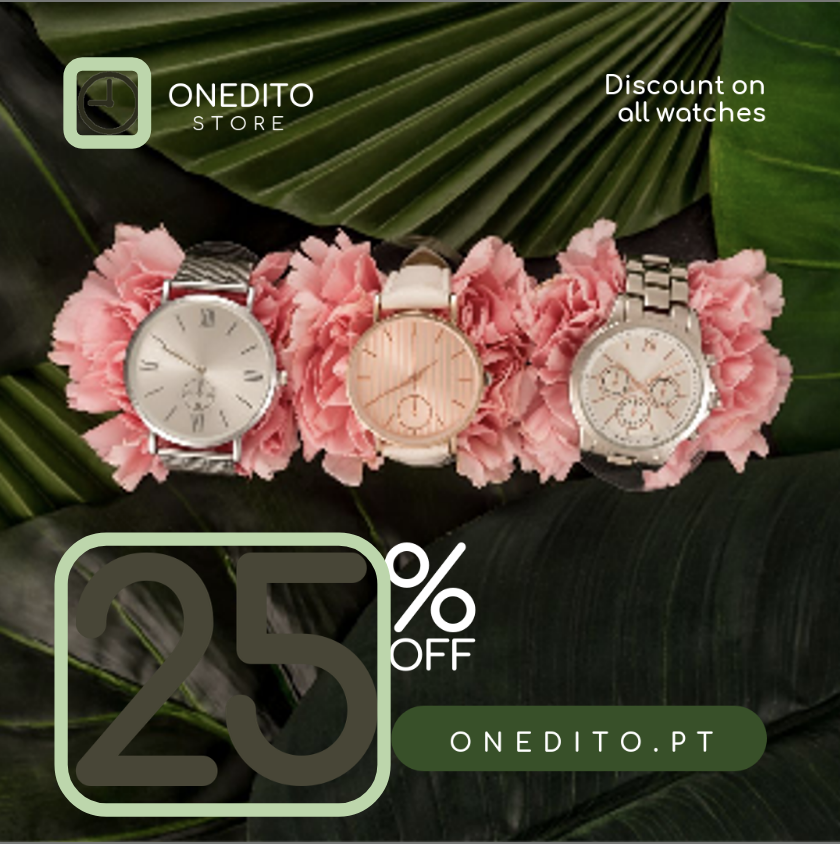} &
\includegraphics[width=0.14\linewidth]{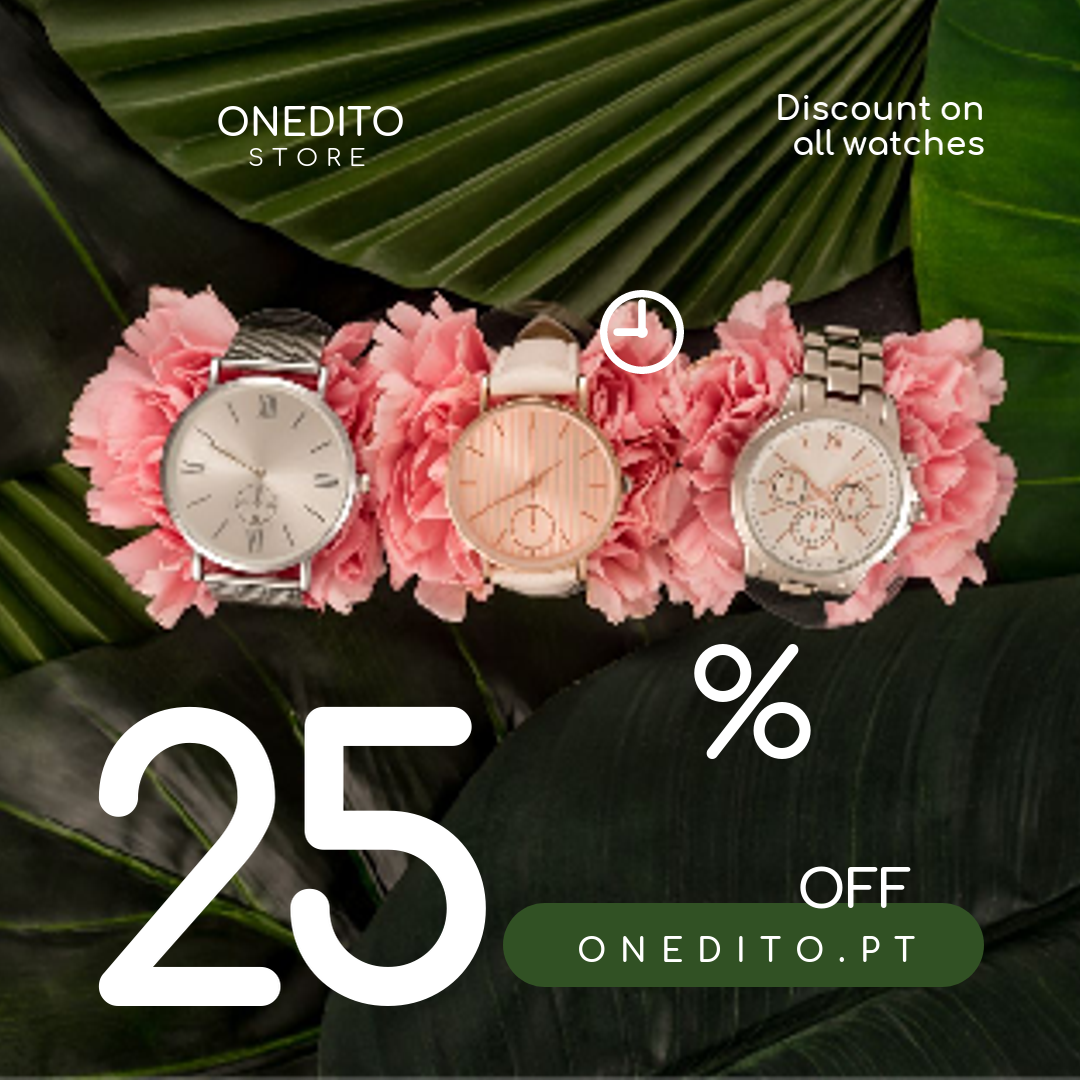} &
\includegraphics[width=0.14\linewidth]{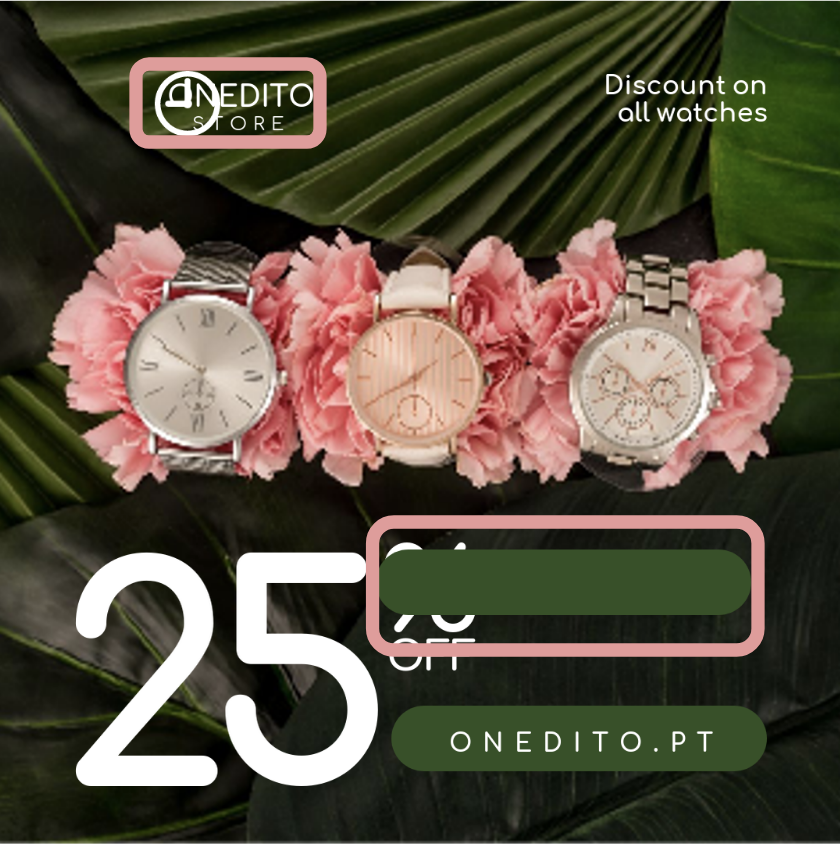} \\[4pt]
\multicolumn{1}{p{0.14\linewidth}}{\centering \small \textit{Original composed layout from Crello.}} & & &
\multicolumn{1}{p{0.14\linewidth}}{\centering \small \textit{Text replaced with a different theme to break thematic consistency.}} &
\multicolumn{1}{p{0.14\linewidth}}{\centering \small \textit{Font attributes like size, color and line height modified to reduce legibility.}} &
\multicolumn{1}{p{0.14\linewidth}}{\centering \small \textit{Element colors adjusted to reduce perceptual contrast.}} &
\multicolumn{1}{p{0.14\linewidth}}{\centering \small \textit{Text boxes and icons displaced to disrupt alignment.}} &
\multicolumn{1}{p{0.14\linewidth}}{\centering \small \textit{Icons, texts and elements overlapped, to induce occlusion.}} \\
\end{tabular}

\caption{
\textbf{Illustration of PRISM perturbations.} The composed layout (left) serves as the reference design. 
Each perturbed layout (right) isolates a violation of a specific design principle: coherence, alignment, readability, contrast, and overlap, while minimally affecting others. 
Descriptions below each image summarize the targeted modification.
Bounding boxes highlight regions where local edits were applied, providing ground-truth supervision for principle-specific localisation.
}
\label{fig:prism_perturbations}
\end{figure*}

Our work makes three key contributions. We
\begin{itemize}
    \item Present \nameemoji{}, a controlled perturbation dataset that isolates principle-specific degradations in designs.
    \item Propose a principled evaluation framework where each principle has a scorer for quantitative assessment and a localiser for spatially grounded feedback.
    \item Demonstrate that our SigLIP-based scorers cleanly separate principle-specific degradations, outperforming GPT-4o in disentanglement 
    (See Fig.\ref{fig:main_radar})
    , and that our instruction-tuned Qwen-2.5-VL localiser aligns most closely with ground-truth annotations for local principles.
\end{itemize}
These advances establish a principled and interpretable pipeline for evaluating and refining visual design.

\section{Related Works}
\label{sec:related}

\textbf{Design Evaluation.} Design evaluation has been extensively studied across domains and development stages. Early heuristic-based methods~\cite{DBLP:journals/ijmms/BauerlyL06,DBLP:conf/doceng/HarringtonNJRT04,ngo2000mathematical,odonovan,DBLP:conf/rcis/ZenV14} struggled to capture semantic design attributes. Subsequent approaches~\cite{DBLP:journals/ijmms/DouZSH19} enhanced evaluation capabilities but lacked comparative scoring mechanisms, motivating siamese-based approaches~\cite{DBLP:journals/tog/ZhaoCL18,goyal2024design}. \citet{DBLP:journals/tvcg/KongJSGCLLZ23} trained a model for aesthetic assessment, while \citet{DBLP:conf/siggraph/TabataYMY19} perturbed layouts to synthesize poor designs for scoring. More recently, MLLM-based methods have focused on design generation~\cite{DBLP:journals/corr/abs-2311-16974}, assessed only limited attributes~\cite{DBLP:journals/corr/abs-2311-16974,gde}, or provided actionable feedback~\cite{duan2024uicrit, sayan2024design}. However, such feedback is often stochastic, redundant, over-suggestive, or lacks contextual awareness - preventing principled, quantifiable and localised assessment of crucial design attributes.

\vspace{3pt}
\noindent\textbf{Design Principles.} Core principles of graphic layout were established long ago and remain widely used ~\cite{odonovan, ODonovan2015DesignScapeDW}. Since then they have been used for guiding layout generation ~\cite{Lok2005ASO, Shi2023IntelligentLG}. Recently, layout-generation methods~\cite{Horita2023RetrievalAugmentedLT, hsu2023posterlayoutnewbenchmarkapproach, vascar, Li2020AttributeConditionedLG, Lin2024InstructLayoutI2} have adopted content based metrics: occulsion, unreadability and graphic metrics including overlay (IoU), non-alignment, using them to iteratively refine layouts toward structurally and visually plausible compositions. Other works also leverage design-related cues for downstream tasks, for example DocEdit-v2~\cite{Suri2024DocEditv2DS} incorporates spatial coherence in multimodal document editing, and hybrid global–local segmentation models use spatial coherence and region–text alignment for grounding~\cite{Xue2024AHL}. Even outside visual design, multimodal topic-modeling introduces metrics to capture visual coherence and topic separability~\cite{GonzalezPizarro2024NeuralMT}. Across these directions, design principles consistently operate as guidance signals for generation and multimodal reasoning, but they remain largely unused as explicit dimensions for assessing design quality.

\vspace{3pt}
\noindent \textbf{Visual grounding with MLLMs.} Recent MLLMs \citep{peng2023kosmos,rasheed2023glamm,chen2023shikra,vistallm,magnet,lai2023lisa,zhao2023bubogpt,wang2023all,you2023ferret} have substantially advanced the integration of visual and linguistic understanding, achieving impressive performance in tasks such as referring expression comprehension and segmentation. These models showcase the growing capability of LLMs to perform fine-grained localisation and complex reasoning over visual content. Approaches such as \citep{peng2023kosmos,chen2023shikra,wang2023all} emphasize language-driven grounding by constructing rich contextual representations. Whereas methods like \citep{lai2023lisa,vistallm,zhao2023bubogpt,rasheed2023glamm,rasheed2023glamm} leverage joint vision-language embeddings to generate segmentation masks. Further, \citep{munasinghe2023pg,meerkat} explore multi-modal grounding by incorporating audio-visual cues for spatial and temporal understanding. Despite these advances, visual grounding within MLLMs remains largely confined to image- and video-based applications and has yet to be extended to specialized domains such as defect localisation in designs.

\section{Principle-Aware Design Evaluation}
To enable interpretable and principle-specific evaluation of visual design quality, 
we introduce a \textbf{principle-aware evaluation framework} built upon a curated dataset of controlled layout perturbations. 
Section~\ref{sec:prism_dataset} presents \textbf{PRISM}, a principle-disentangled dataset derived from professional designs that isolates key visual composition principles through targeted modifications. 
Section~\ref{sec:scorer_localiser} then describes the \textbf{scorer} and \textbf{localiser} modules that leverage PRISM to quantitatively and spatially assess design violations.

\begin{figure*}[t]
     \centering
     \includegraphics[width=\linewidth, trim={0.2cm 4cm 1cm 3cm},clip]{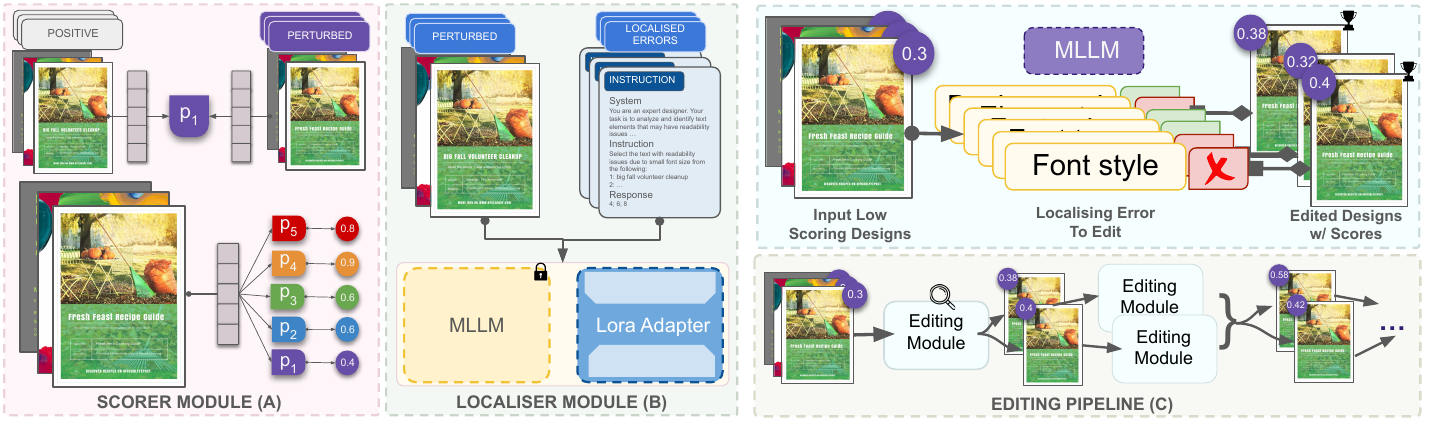}
     \caption{Using \nameemoji{}, we build two modules for principle-aware design evaluation: (A) a \textbf{scorer} that quantifies quality across five design principles using controlled perturbations, and (B) a \textbf{localiser} that identifies regions responsible for readability or contrast degradations. A \textbf{demonstrative editing pipeline} (C) operates on the coherence module, where GPT-4o serves as a prompting-based localiser to identify global inconsistencies and propose edits. The pipeline employs a \textbf{beam search} strategy, generating candidate edits from the localiser and selecting the top two (beam size = 2) using the scorer. Together, these components establish a foundation for interpretable, end-to-end design assessment and refinement.}
     \label{fig:method}
 \end{figure*}
\subsection{\nameemoji{}: Principle-Disentangled Dataset}
\label{sec:prism_dataset}
To enable interpretable and principle-specific evaluation of visual layouts, we introduce \textbf{PRISM} (\textbf{PR}inciple-aware, \textbf{I}nterpretable, and \textbf{S}tructure-guided Design \textbf{M}odifications), a curated dataset that disentangles five fundamental principles of design quality: \textit{coherence}, \textit{readability}, \textit{contrast}, \textit{alignment}, and \textit{overlap} (See Fig. \ref{fig:pull_figure}).  
PRISM is derived from the Crello corpus ~\cite{canvasvae} of professional designs, which provides structured metadata describing each design’s geometry, text attributes, color palette, and so on. 
This metadata enables controlled perturbations of individual design components while preserving other aspects of composition.

\vspace{3pt}
\noindent\textbf{Principle definitions.}  
We define the principles as follows:  
\textit{\textbf{Coherence}} measures semantic consistency between textual and visual elements;  \textit{\textbf{Readability}} measures typographic clarity and legibility;  
\textit{\textbf{Contrast}} measures perceptual distinctness between overlapping or adjacent regions; \textit{\textbf{Alignment}} measures geometric regularity of text and element placement; and 
\textit{\textbf{Overlap}} measures unintended occlusions between visual components.  
These definitions serve as the organizing factors guiding the perturbations ~\cite{tacit,huang2023survey,lidwell2010universal}.

\vspace{3pt}
\noindent\textbf{Perturbation methodology.}  
For each principle \(p \in \mathcal{P}\), we generate a perturbed layout \(x'_p = x + \Delta_p\), 
where \(\Delta_p\) introduces a targeted violation of \(p\) while keeping other principles approximately invariant.  
\textbf{Global} principles: \textit{coherence} and \textit{alignment} require semantic or structural modifications, whereas  
\textbf{local} principles: \textit{readability}, \textit{contrast}, and \textit{overlap} involve perceptual or spatial adjustments.  
Across all principles, GPT-4o–based \cite{gpt4o} prompting is used to modify textual content, replace icons or background imagery, and suggest RGB color values for altered text and elements. The prompt does not include the poster as a whole but only the required details for edits (Supp.A).   
In parallel, geometric and typographic attributes such as text-box coordinates, font size, etc are adjusted directly through the Crello metadata, scaled relative to their original values to ensure realistic, layout-consistent perturbations. For local principles, we additionally record the precise set of modified regions and corresponding changes, providing structured supervision that can be used to evaluate or guide localisation-based editing techniques (See Section~\ref{sec:localised_reasoning}). 

\noindent\textbf{Example.}  
Figure~\ref{fig:prism_perturbations} illustrates representative perturbations.  
In the \textit{coherence} example, the text promotes a ``dolphin show’’ while the icons and background depict wristwatches, violating thematic consistency.  
In the \textit{readability} example, we intentionally reduced the font size for the word ``onedito'', which reduces readability.

Although complete independence among principles is inherently challenging; each perturbation is constructed to emphasize a single factor with minimal collateral impact on others. 
The resulting dataset comprises paired composed and perturbed layouts \(\{(x, x'_p)\}\), enabling quantitative and spatially grounded evaluation of design quality. 
\textbf{PRISM} is derived from the Crello corpus, which provides approximately 20K training and 2K validation professional designs. 
Each principle is perturbed independently over this split, yielding about \textbf{20K training} and \textbf{2K validation} per principle, and totaling roughly \textbf{100K training} and \textbf{10K validation} layout pairs across all five principles.


\subsection{Scorer and Localiser Modules}
\label{sec:scorer_localiser}
Building on PRISM, we develop two complementary modules for design evaluation. 
The \textbf{binary scorers} quantitatively assess whether each principle is satisfied in a given layout, 
while the \textbf{localiser} provides spatially grounded feedback identifying regions responsible for the violation. 
Together, these components enable fine-grained, interpretable evaluation of visual layouts and serve as inputs for the editing pipeline described in Section~\ref{sec:editing}.

\subsubsection{Principle-Aware Scorer}
\label{sec:scorer}
We define a collection of \textit{principle-aware grounding functions} 
$\mathcal{G} = \{ g_p \}_{p \in \mathcal{P}}$
that estimate the likelihood that a visual layout satisfies specific design principles, such as 
\textit{readability}, \textit{contrast}, \textit{alignment}, \textit{overlap} and \textit{coherence}.
Each grounding function $g_p$ operates over a design representation $\phi(x)$ 
and outputs a probability 
$P(p\,|\,x) = g_p(\phi(x))$, 
quantifying adherence to principle $p$.
Together, these functions provide an interpretable decomposition of overall design quality, 
mapping a layout’s multimodal embedding space into principle-specific probabilities.
This formulation enables a structured and transparent evaluation framework 
that moves beyond a single, opaque score
toward measurable reasoning aligned with human design intuition.

\vspace{3pt}
\noindent \textbf{Design-Aware Fine-tuning.}
\label{sec:design_finetuning}
We begin by adapting a pretrained multi-modal encoder to become \textit{design-aware}, ensuring its visual representations capture structural and compositional cues relevant to layout evaluation. 
At this stage, only the vision encoder is updated, while the text encoder remains frozen.
The model is trained on multi-modal document-caption pairs, where the captions convey the overall design intent of the document. Through this adaptation, the encoder learns layout embeddings that capture spatial arrangement, visual balance, and typographic hierarchy. The resulting design-aware vision encoder produces fixed embeddings $\phi(x)$ for all scorers, ensuring consistent and semantically grounded feature representations across principles.

\vspace{3pt}
\noindent \textbf{PRISM-Scorer.} We train independent binary classifiers on top of the frozen design-aware embeddings to estimate whether each design principle $p \in \mathcal{P}$ is satisfied. 
Given a layout $x$, the encoder produces $\phi(x)$, and the corresponding scorer predicts: 
\begin{equation}
    g_p(x) = \sigma(\phi(x)^{\top} W_p),
\end{equation}
where $W_p \in \mathbb{R}^{d}$ are the learnable weights for principle $p$, 
and $\sigma(\cdot)$ denotes the sigmoid activation. 
Training uses paired composed and perturbed examples, 
where each perturbation selectively violates a single design principle. 
Each scorer is optimized using binary supervision and learns to assign higher probabilities to principle-aligned layout designs 
and lower probabilities to those designs that are at odds with key design principles.
The resulting outputs $g_p(x) \in [0,1]$ form probabilistic, interpretable scores 
indicating the degree of adherence to each principle, as illustrated in Fig.~\ref{fig:method}(A). Collectively, the set of principle-specific functions
$\mathcal{G}(x) = \{ g_p(x) \}_{p \in \mathcal{P}}$
constitutes a structured, interpretable evaluation space that decomposes overall design quality 
into measurable dimensions aligned with human reasoning.

\subsubsection{Principle-specific Localisation of Errors}
While the binary scorers provide quantitative assessments of whether a design adheres to each principle, they do not reveal \textit{where} violations occur. To enable spatially and semantically grounded interpretability, we introduce a principle-specific localisation module that evaluates beyond scalar scores. 
This module identifies the precise visual or textual elements responsible for violating a given design principle, 
bridging quantitative scoring with qualitative explanation. 
Localisation is modeled at two complementary levels:
(\textit{i}) a \textbf{localised} component that detects fine-grained perceptual issues such as readability, contrast, and overlap, and
(\textit{ii}) a \textbf{global} component that reasons about higher-level semantic properties such as coherence and alignment.

\vspace{3pt}
\label{sec:localised_reasoning}
\noindent\textbf{Localised Reasoning for Perceptual Principles.} To localise design violations and provide interpretable feedback for perceptual principles such as 
\textit{readability}, \textit{contrast}, and \textit{overlap}, 
we fine-tune a vision-language model using structured instruction prompts derived from localised perturbations in PRISM. 
This module identifies the specific \textit{elements} within a layout that fail to satisfy a given principle.

During training, each example is formulated as an instruction-based multiple-choice task with explicit candidate options from the metadata. 
For \textit{readability}, the input includes the layout image and a list of text elements, each associated with a unique element identifier and its textual content. 
For \textit{contrast} and \textit{overlap}, the input contains pairs of elements that occupy intersecting regions in the design space, 
also represented by their corresponding IDs. 
The instruction then prompts the model to select which elements or element pairs violate the specified principle, 
for example: ``Select the text with readability issues due to small font size.''
To establish consistent reasoning behavior, we include a system prompt that defines the model’s role and context at the start of every interaction, shown in Fig. ~\ref{fig:method}(B). 

Given an image--instruction pair $(x, c_p)$ and a finite candidate set $\mathcal{O} = \{o_1, o_2, \dots, o_n\}$, 
where each $o_i$ denotes an element ID and its associated content, 
the model predicts a subset $\hat{\mathcal{O}}_p \subseteq \mathcal{O}$ corresponding to the violating elements:
\begin{equation}
    \hat{\mathcal{O}}_p = \text{Model}(x, c_p, \mathcal{O}).
\end{equation}
Training minimizes the cross-entropy loss between the predicted and ground-truth selections of violating element IDs.  

During inference, the fine-tuned model outputs the list of IDs of the elements that violate the queried principle, 
without generating free-form text. 
This design choice constrains the output space to a discrete set of candidates, 
improving reliability and reproducibility over generative responses. 

\vspace{3pt}
\noindent \textbf{Global Reasoning for Semantic Principles.}
While perceptual principles such as \textit{readability}, \textit{contrast}, and \textit{overlap} can be localised to specific regions, global principles like \textit{coherence} and \textit{alignment} require reasoning over the entire layout.
PRISM provides structured perturbations for these principles, but a definitive ground truth is not available since multiple valid revisions can achieve coherence or alignment.
We therefore use a prompt-based evaluation that checks whether major visual components align with the layout's inferred theme and structure, enabling holistic assessment of semantic and organizational fidelity.

For instance, while examining \textit{coherence}, the model first infers the design's overarching theme (e.g., ``Kid's Birthday Party'') from its textual content and derives representative visual expectations for each component, such as a playful background, festive icons, and bold, cheerful typography. Each component (background, icons, font style, etc.) is then queried using structured prompts to verify alignment with these expectations.
This procedure identifies components that visually differ from the inferred theme.

An analogous formulation is applied to \textit{alignment}, where the model verifies whether elements such as text boxes or icons deviate from the underlying structural grid.
The resulting qualitative judgments provide interpretable indicators of global principle violations; not all require correction, and resolving one often restores multiple inconsistencies.

\begin{figure}[t!]
     \centering
     \includegraphics[width=0.9\linewidth, trim={0cm 0cm 0cm 0cm},clip]{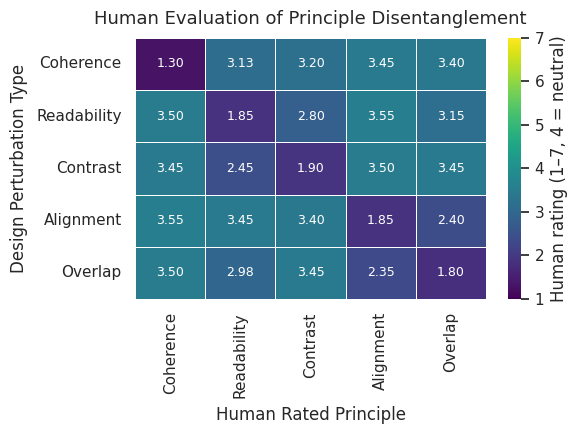}
     \caption{\textbf{Human evaluation of principle disentanglement.} 
Average ratings (1–7, 4 = neutral) across 10 poster pairs per principle show strong diagonal responses, confirming that PRISM perturbations primarily affect their intended design principles.}
     \label{fig:human_eval}
     \vspace{-0.2in}
 \end{figure}
\section{Experimental Setup}
\label{sec:Experiments}
Our experiments evaluate whether PRISM produces clean, principle-specific degradations and examine how effectively our scorer–localiser framework and contemporary MLLMs detect and respond to these controlled violations.

\subsection{Validating Principle Disentanglement}
To assess whether PRISM perturbations successfully isolate individual design principles, we conduct a controlled human study with 15 participants. Each participant is shown 10 paired examples, where each pair contains a professionally designed poster and its perturbed variant, without indicating which is which. For every pair, three independent participants rate which design better satisfies each of the five principles (\textit{coherence}, \textit{readability}, \textit{contrast}, \textit{alignment}, and \textit{overlap}) on a 1–7 rating scale, where 1 indicates the original is much better, 7 indicates the perturbed is much better, and 4 indicates no clear preference. We get the average of three participants for each pair and calculate the mean across all pairs. This measures how selectively each perturbation affects its intended principle.


\begin{figure}[t!]
     \centering
     \includegraphics[width=0.9\linewidth, trim={0cm 0cm 14cm 0cm},clip]{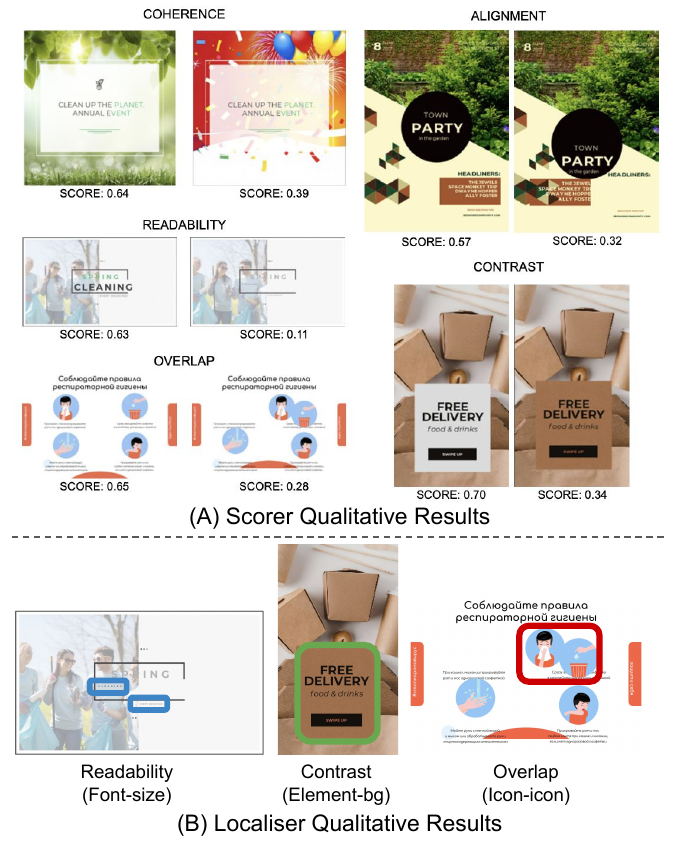}
     \caption{\textbf{Qualitative results.} (A)\textbf{ PRISM-Scorer outputs.} The left image in each pair is the original layout and the right shows its PRISM perturbation, with scores indicating the predicted principle-specific degradation. (B) \textbf{Localiser outputs} from the Qwen-2.5-VL instruct-tuned model. The highlighted text specifies the targeted principle and the sub-aspect being queried. For example, in readability we prompt the model to attend specifically to small font-size regions that impair clarity.}
     \label{fig:qualitative}
     \vspace{-0.2in}
 \end{figure}

\subsection{Training of Principle-Aware Scorers}
Each principle-specific scorer is trained on a design-aware SigLIP-v2 vision encoder \cite{siglipv2}, following the setup in Section~\ref{sec:scorer}.
The encoder is adapted using the training splits of the Creatidesign \cite{zhang2025creatidesignunifiedmulticonditionaldiffusion} and Crello \cite{canvasvae} datasets, containing 400K and 20K layout–caption pairs.
Captions describing each poster’s theme and composition are generated with GPT-4o \cite{gpt4o}.
Only the vision encoder is updated in this stage, enabling it to capture structural, spatial, and typographic cues relevant to layout understanding.

Using the frozen encoder, we train five binary classifiers, one for each design principle, to distinguish between composed and perturbed layouts. Each scorer is trained and evaluated on balanced sets comprising composed layouts from the Crello dataset and their corresponding perturbations from the PRISM dataset, using 40K images for training and 4K for testing.
A 90–10 split is used for validation within the training set.
Training employs binary cross-entropy loss, with model selection based on validation AUC, and early stopping is applied after seven consecutive non-improving epochs.
We report Precision, Recall, F1, and AUC to quantify scorer performance across principles.

\subsection{Model Sensitivity to PRISM Perturbations}
\label{sec:model_sensitivity}
To assess how well different models capture principle-specific degradations, we evaluate four systems on PRISM: GPT-4o, GPT-4o-mini, Qwen-2.5-VL (7B) \cite{qwen2_5}, and our principle-aware PRISM-scorer.
For each design principle, we sample 200 examples (100 composed–perturbed pairs).
Each model is queried to provide ratings for all five principles for every poster.
We then compute the mean score for original and perturbed designs separately, and take their absolute difference ($\Delta$) as a measure of sensitivity to that perturbation.
A higher $\Delta$ indicates stronger discrimination between high- and low-quality designs for that principle. Refer to Supp. for prompt details.

\subsection{Training of Localiser}
\label{sec:localiser_training}
We instruct-tune Qwen-2.5-VL (7B) using the localised annotations from the PRISM dataset, following the formulation described in Section~\ref{sec:localised_reasoning}. 
This experiment is conducted only for the \textit{readability}, \textit{contrast}, and \textit{overlap} principles, as these can be spatially localised to specific text or visual elements within the layout. 
For each layout, the model predicts a set $P$ of elements that violate the targeted principle, which is compared against the ground-truth set $G$ from PRISM. 
True positives ($TP$) correspond to elements present in both $P$ and $G$, false positives ($FP$) to elements in $P$ but not in $G$, and false negatives ($FN$) to elements in $G$ but missing from $P$. 
We compute Precision, Recall, and F1-score based on these counts, and Intersection-over-Union (IoU) using the sets to evaluate localization accuracy. 
To ensure consistency across models, we use the same system prompt for all baselines and instruct-tuned variants. 
Instruct-tuning is performed with one LoRA adapter \cite{hu2021loralowrankadaptationlarge} per principle, allowing independent specialization for \textit{readability}, \textit{contrast}, and \textit{overlap}. 
We compare the instruct-tuned Qwen-2.5-VL model against three baselines: the pretrained Qwen-2.5-VL (base) model, a few-shot (3-shot) prompted version using the same instruction template, GPT-4o-mini, and GPT-4o queried with equivalent localised prompts. 
This comparison evaluates how PRISM perturbations enhance localisation.


\begin{table}[t!]
\setlength{\tabcolsep}{6pt}
\renewcommand{\arraystretch}{0.7}
\resizebox{\linewidth}{!}{\begin{tabular}{l|cccc}
\toprule
\textbf{PRISM-scorer} & \textbf{Precision} $\uparrow$ & \textbf{Recall} $\uparrow$ & \textbf{F1-score} $\uparrow$     & \textbf{AUC} $\uparrow$    \\ \midrule
Coherence    & 0.7243    & 0.7279 & 0.7262 & 0.7983 \\
Readability  & 0.7336    & 0.7323 & 0.7329 & 0.8064 \\
Contrast     & 0.7552    & 0.7486 & 0.7519 & 0.8291 \\
Alignment    & 0.7648    & 0.7591 & 0.7619 & 0.8321 \\
Overlap      & 0.7175    & 0.7232 & 0.7203 & 0.7973 \\
\bottomrule
\end{tabular}}
\caption{\textbf{Performance of PRISM binary scorers.} 
Results on the held-out validation set showing Precision, Recall, F1, and AUC for each design principle.}
\label{tab:scorer_results}
\end{table}
\begin{figure*}[t]
     \centering
     \includegraphics[width=\linewidth, trim={0cm 0cm 0cm 0cm},clip]{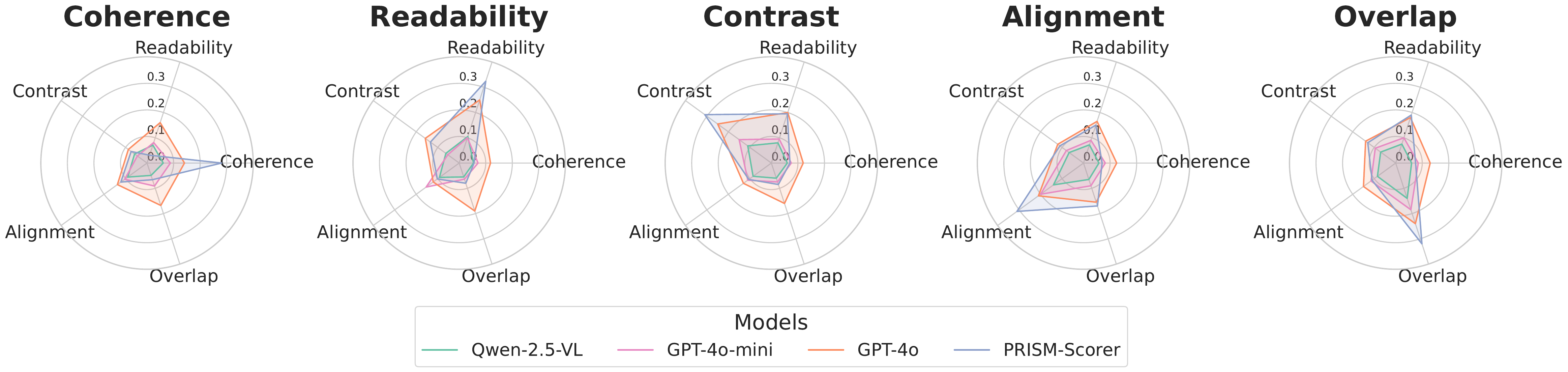}
     \caption{\textbf{Model sensitivity to PRISM perturbations.} 
Radar plots show average rating differences (\(\Delta\)) between composed and perturbed posters across principles. 
Our PRISM scorer demonstrates the strongest and most principle-specific sensitivity.}
     \label{fig:radar_results}
 \end{figure*}
\begin{table*}[t]
\centering
\resizebox{\linewidth}{!}{
\begin{tabular}{lccccccccccccccc}
\toprule
\multirow{2}{*}{\textbf{Model}} &  & \multicolumn{4}{c}{\textbf{Readability}}      &  & \multicolumn{4}{c}{\textbf{Contrast}}  &  & \multicolumn{4}{c}{\textbf{Overlap}}   \\ 
\cmidrule{3-6} \cmidrule{8-11} \cmidrule{13-16} 
 &  & \bf IoU $\uparrow$    & \bf Precision $\uparrow$ & \bf Recall $\uparrow$ & \bf F1-score $\uparrow$     &  & \bf IoU $\uparrow$ & \bf Precision $\uparrow$ & \bf Recall $\uparrow$ & \bf F1-score $\uparrow$ &  & \bf IoU $\uparrow$ & \bf Precision $\uparrow$ & \bf Recall $\uparrow$ & \bf F1-score $\uparrow$ \\ 
\midrule
Qwen$_{\text{Base}}$       &  & 0.3645 & 0.4091 & 0.5478 & 0.4312 &  & 0.2952 & 0.3314 & 0.4725 & 0.3804 &  & 0.3215 & 0.3411 & 0.4513 & 0.3770 \\
Qwen$_{\text{Prompted}}$   &  & 0.3603 & 0.4303 & 0.4782 & 0.4147 &  & 0.3186 & 0.3519 & 0.4884 & 0.4049 &  & 0.3297 & 0.3520 & 0.4728 & 0.3984 \\
GPT-4o-mini              &  & 0.4028 & 0.4387 & 0.6127 & 0.4718 &  & 0.3456 & 0.3924 & 0.5311 & 0.4478 &  & 0.3729 & 0.4012 & 0.5238 & 0.4526 \\
GPT-4o                   &  & 0.5532 & 0.6433 & 0.6228 & 0.6037 &  & 0.5164 & 0.5832 & 0.6215 & 0.6017 &  & 0.5417 & 0.6112 & 0.6354 & 0.6210 \\
\rowcolor{cyan!10} \bf Qwen$_{\text{Expert}}$ (Ours) &  & \bf 0.7833 & \bf 0.8015 & \bf 0.8132 & \bf 0.7998 &  & \bf 0.6761 & \bf 0.7024 & \bf 0.7452 & \bf 0.7196 &  & \bf 0.7328 & \bf 0.7553 & \bf 0.7914 & \bf 0.7730 \\ \midrule
$\bf \textcolor{blue!60}{\Delta_{\text{\bf Ours} - \text{\bf GPT-4o}}}$ &  & \textcolor{blue!60}{\bf 0.230$\uparrow$} & \textcolor{blue!60}{\bf 0.158$\uparrow$} & \textcolor{blue!60}{\bf 0.190$\uparrow$} & \textcolor{blue!60}{\bf 0.196$\uparrow$} &  & \textcolor{blue!60}{\bf 0.160$\uparrow$} & \textcolor{blue!60}{\bf 0.119$\uparrow$} & \textcolor{blue!60}{\bf 0.124$\uparrow$} & \textcolor{blue!60}{\bf 0.118$\uparrow$} &  & \textcolor{blue!60}{\bf 0.191$\uparrow$} & \textcolor{blue!60}{\bf 0.144$\uparrow$} & \textcolor{blue!60}{\bf 0.156$\uparrow$} & \textcolor{blue!60}{\bf 0.152$\uparrow$} \\
\bottomrule
\end{tabular}
}
\caption{\textbf{Localisation performance across principles.} 
Fine-tuning Qwen-2.5-VL with PRISM annotations leads to substantial improvements in IoU and F1 across readability, contrast, and overlap compared to base, prompted, GPT-4o-mini and GPT-4o.}
\label{tab:localiser_results}
\end{table*}

\section{Main Results}
All analyses use the PRISM held-out validation split to assess responses to principle-specific degradations.

\subsection{Principle Disentanglement: Human Evaluation}
Figure~\ref{fig:human_eval} displays average human ratings across perturbation types and design principles. Diagonal scores are below the neutral midpoint of 4, indicating a strong preference for original designs over perturbed ones. For instance, coherence perturbations averaged 1.3 (coherence score), confirming their impact on thematic consistency. Off-diagonal ratings remained near 4, showing minimal effects on unrelated principles. We also observe a correlation between readability and contrast, and between alignment and overlap, which reflects the natural perceptual coupling among these visual attributes. These results show that PRISM perturbations introduce controlled, interpretable degradations that are perceptually distinct across design principles.

\subsection{PRISM Scorer: Performance \& Sensitivity}
We first evaluate the PRISM binary scorers on the held-out validation set to assess their reliability across design principles. Qualitative examples illustrating these scorer responses are described in Figure~\ref{fig:qualitative}(A).
As shown in Table~\ref{tab:scorer_results}, the scorers achieve moderate but consistent performance, with F1-scores ranging between 0.72 and 0.76 and AUC values between 0.80 and 0.82. 
While these results do not reflect perfect classification accuracy, the objective of the scorers is to achieve stable and interpretable sensitivity to principle-specific degradations, rather than absolute predictive performance. 
This sensitivity makes the scorers reliable quantitative indicators of principle-specific degradations.

Figure~\ref{fig:radar_results} visualizes model sensitivities across principles using radar plots.
Each plot corresponds to one perturbation type and shows the absolute change in model ratings between composed and perturbed posters.
The PRISM scorers exhibit the most principle-aligned behavior, responding strongly along the targeted dimension while remaining stable elsewhere. As each scorer is only trained on one specific principle, all other principles represent out-of-domain data. 
GPT-4o captures overall degradation trends but does not disentangle individual principles, and GPT-4o-mini and Qwen-2.5-VL show weaker, noisier responses.
These findings indicate that the PRISM scorers capture fine-grained degradations and provide a basis for principle-aware evaluation of visual design quality. (Refer to Supp for additional OOD experiments).

\subsection{PRISM: Localised Error Detection}
\label{sec:localiser_results}
Table~\ref{tab:localiser_results} presents the localisation performance across the \textit{readability}, \textit{contrast}, and \textit{overlap} principles. 
Instruct-tuning Qwen-2.5-VL with PRISM annotations substantially improves performance over the base, prompted, and GPT-based models, achieving the highest IoU and F1-scores across all principles. 
The instruct-tuned model achieves an average F1 score of 0.80 for \textit{readability}, 0.72 for \textit{contrast}, and 0.77 for \textit{overlap}, demonstrating its ability to identify localised design violations accurately substantially outperforming GPT-4o which demonstrates moderate localisation ability but tends to produce less precise predictions. GPT-4o-mini and prompted Qwen-2.5-VL exhibit weaker and noisier trends. 
These results show that principle-specific fine-tuning with PRISM supervision leads to more interpretable and spatially grounded reasoning for detecting localised degradations. We also provide qualitative localiser outputs for these principles in Figure~\ref{fig:qualitative}(B).


\begin{figure}[t!]
\centering
\renewcommand{\arraystretch}{1.05}
\setlength{\tabcolsep}{4pt}

\begin{tabular}{lllllll}
\multicolumn{1}{c}{\textbf{Score:0.19}} & &
\multicolumn{1}{c}{\hspace{-14pt}\textbf{Score: 0.62}} & &
\multicolumn{1}{c}{\hspace{-14pt}\textbf{Score: 0.66}}\\
\includegraphics[width=0.27\linewidth]{figures/examples/coherence.png} & 
\hspace{-7pt}\includegraphics[width=0.07\linewidth]{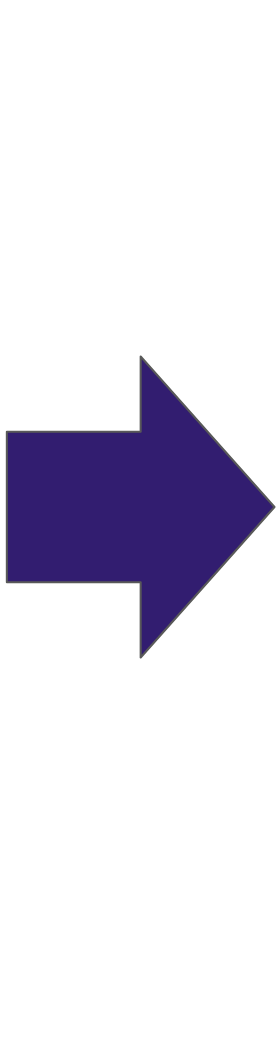} &
\hspace{-7pt}\includegraphics[width=0.27\linewidth]{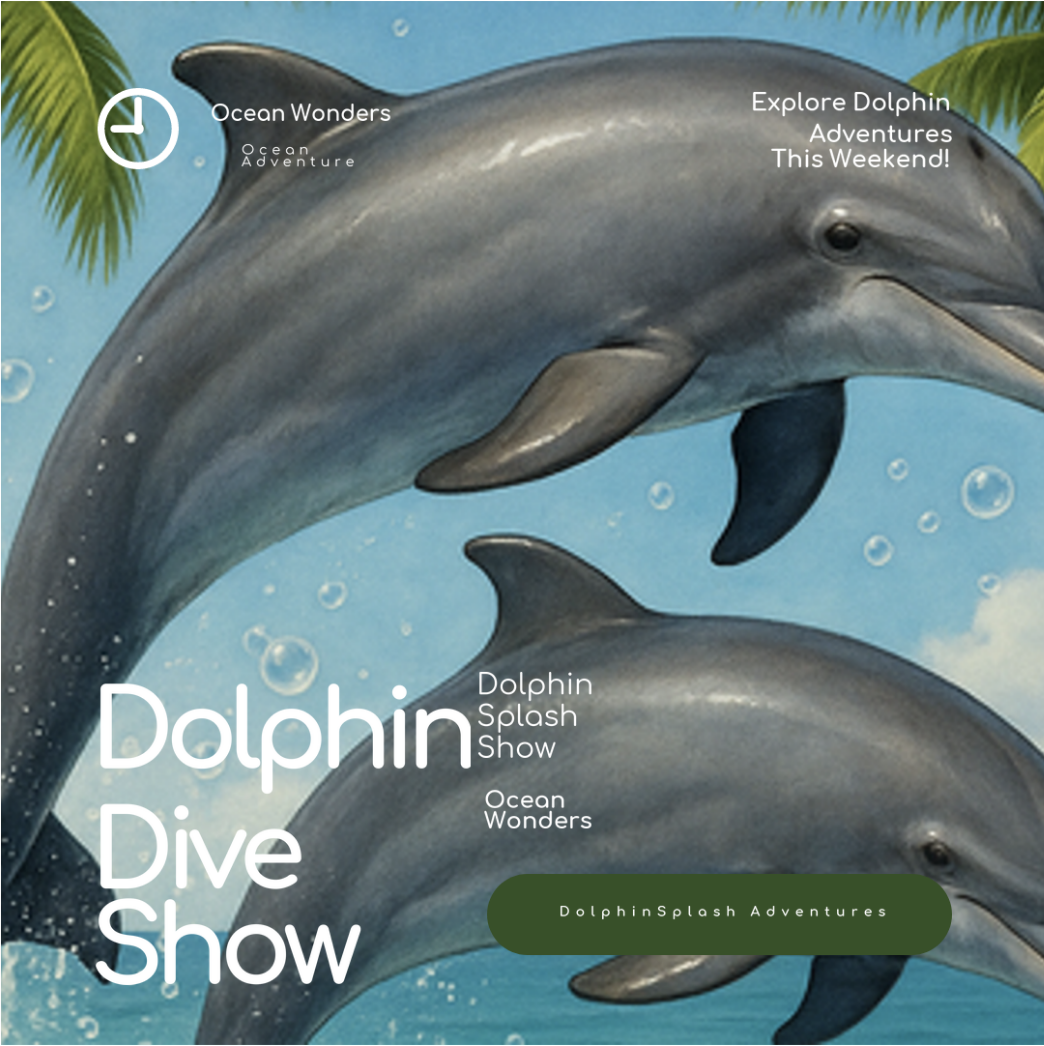} &
\hspace{-14pt}\includegraphics[width=0.07\linewidth]{figures/editing/arrow.png} &
\hspace{-7pt}\includegraphics[width=0.27\linewidth]{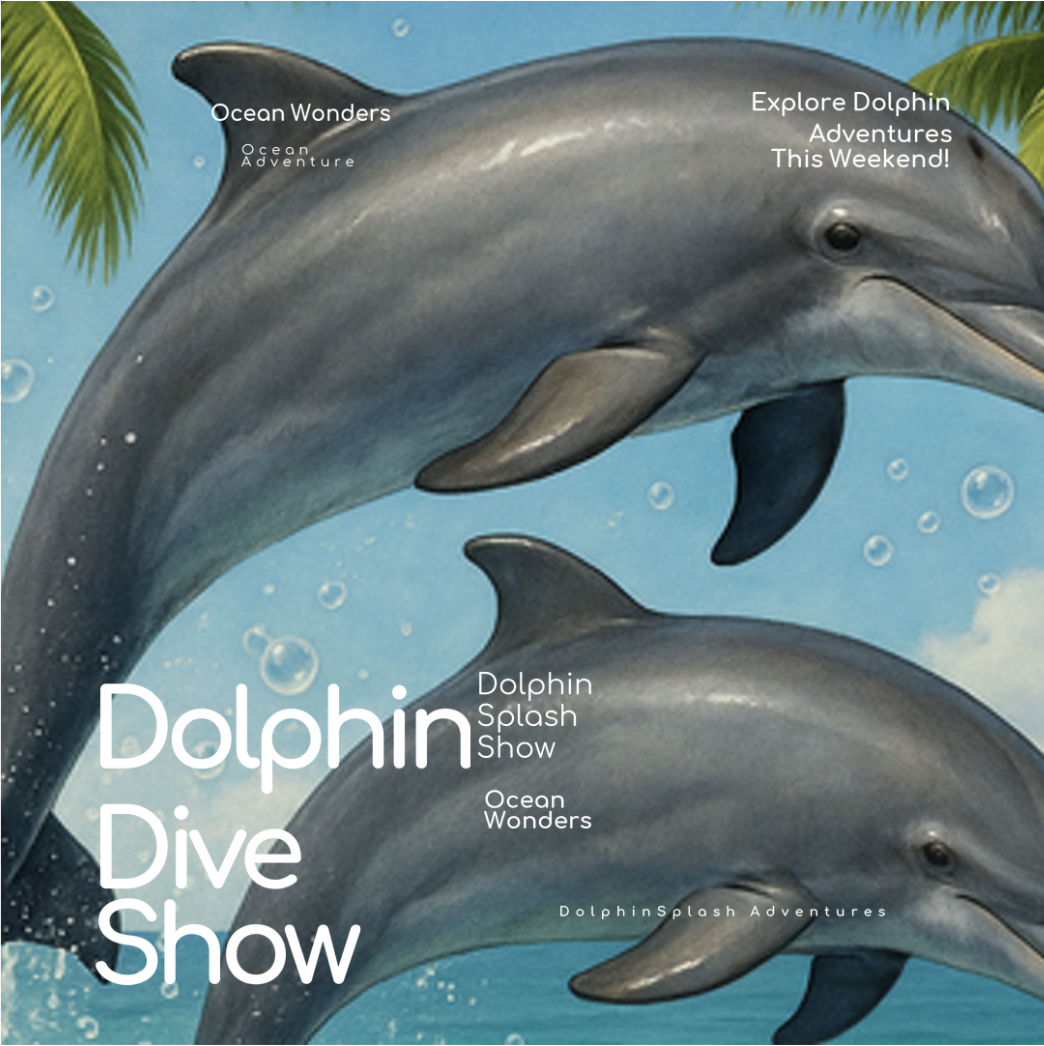}\\[4pt]
\multicolumn{1}{p{0.27\linewidth}}{\centering \small (a) Incoherent Design} & &
\multicolumn{1}{p{0.27\linewidth}}{\hspace{-14pt}\centering \small (b) Background \hspace{-7pt}replaced} & &
\multicolumn{1}{p{0.27\linewidth}}{\hspace{-14pt}\centering \small (c) Irrelevant Icon\hspace{-7pt}removed} \\
\end{tabular}

\caption{\textbf{Demonstrative coherence editing pipeline.} 
Starting from an incoherent design (a) from the PRISM dataset, the model iteratively refines the layout guided by the coherence scorer and localiser. 
The first edit (b) replaces the mismatched background with a theme-consistent image, while the second edit (c) removes the irrelevant watch icon. 
The coherence score increases from 0.19~$\rightarrow$~0.62~$\rightarrow$~0.66, reflecting improved thematic consistency.}
\label{fig:editing}
\vspace{-0.2in}
\end{figure}
\section{Demonstrative Editing Pipeline}
\label{sec:editing}
To illustrate a possible utility of our framework, we implement a demonstrative editing pipeline for the coherence principle using a beam search-based strategy \cite{shi2024thorough}. 
The pipeline uses feedback from the scorer and localisation modules to propose and evaluate principle-aware refinements, as introduced in Fig.~\ref{fig:method}(C). 
At each iteration, the localisation module identifies elements contributing to incoherence and helping generate candidate edits through prompt-based reasoning. 
Each candidate layout addresses one of the identified inconsistencies, yielding multiple alternative paths for editing. 
The coherence scorer re-evaluates these candidates, and the top-$k$ candidates (with a beam size $k=2$) achieving the highest coherence scores are retained for subsequent iterations. 
Unlike direct modifications from MLLMs, this pipeline preserves the underlying layout structure and retains the design’s editability throughout the refinement process. 
See Fig.~\ref{fig:editing} for an illustrative example, demonstrating how interpretable feedback enables targeted, principle-specific refinements in visual design.

\section{Conclusions and Future Work}
\label{sec:discussion}

\noindent \textbf{Conclusions.} We introduce \nameemoji{}, a principle-disentangled benchmark and a multi-scale evaluation framework for interpretable reasoning of visual designs. PRISM enables consistent comparison of disentanglement skills by introducing a principle-aware perturbed dataset. We combine lightweight scorers and instruction-tuned models for localising errors, which can be used to propose targeted edits. These components lay a foundation for future multimodal systems that can diagnose, explain, and improve visual layouts aligned with human designers. 


\vspace{3pt}

\noindent \textbf{Future Work.}
While PRISM offers controlled principle-specific perturbations, there are several opportunities for improvement. First, the assumption of strict principle independence is inexact (for instance, readability and contrast can influence one another). A possible avenue for future work is to model such interactions more explicitly. Second, current coherence assessments rely on prompt-based judgments that vary across generative models; greater robustness may be achieved through multiple annotators, more stable prompting strategies, or the incorporation of theme-level annotations. Third, the editing pipeline could be extended to support vector-level edits. Finally, the scorer and localiser can double as reward functions for RLHF- or DPO-style \cite{rlhf, rafailov2024directpreferenceoptimizationlanguage} training, allowing models to learn scoring, localisation, and editing, holistically.

{
    \small
    \bibliographystyle{ieeenat_fullname}
    \bibliography{references}
}

\clearpage
\appendix
\setcounter{page}{1}
\maketitlesupplementary
\setcounter{figure}{8}   
\setcounter{table}{2}    

\noindent\textbf{The supplementary is organised as follows:}\\
\ref{sec:perturbation_details} PRISM Perturbations\\
\ref{sec:sensitivity_prompts} Prompting Setup for Model Sensitivity\\
\ref{sec:backbone} Backbone Ablation\\
\ref{sec:localised_details} Localised Error Detection Details\\
\ref{sec:OOD} Scorer Out-of-Domain Generalization\\
\ref{sec:more_examples} Additional Examples from Editing Pipeline\\


\section{PRISM Perturbations}
\label{sec:perturbation_details}
In this section we provide additional details about how each perturbation in PRISM is constructed. The main paper introduces the framework and motivation in Section 3.1.
Below, we describe the procedures used to isolate each design principle while keeping all other aspects of the poster unchanged so that each variant reflects a targeted violation of only one dimension. For the perturbations that provide localized supervision (readability, contrast, and overlap), we also record the elements involved so they can be used during instruction-tuning (Section 3.2.2).

\subsection{Coherence}

For coherence perturbation, we reassign each poster a new semantic theme sampled from a different category. We apply either a text-based or element-based modification selected at random. In the text perturbation, all textual content is rewritten using a few-shot LLM prompt so that the new text reflects the target theme while maintaining similar length. Only the text in the metadata is modified, thus preserving attributes such as font style, boldness, alignment, spacing, and color. For element modification, we identify visually prominent components using a vision-language query over the composed layout and check their relevance to the original theme using a lightweight LLM-based relevance classifier. We remove or replace elements determined to be theme-specific, and important background regions are regenerated using an image editor conditioned on the new theme while maintaining the color family. The new design preserves structure but conveys a different semantic theme.

\subsection{Readability}

Readability perturbation reduces the legibility of text while keeping the overall layout intact. For each poster, we randomly select a subset of text elements and apply one or more readability changes. These include shrinking text that is unusually large, adjusting the line height of multi-line text blocks so that spacing becomes compressed, and shifting of text color toward the dominant background color behind it. Each poster receives only a subset of these modifications, and the perturbations applied to each text block are recorded so that the dataset includes explicit supervision about how readability was degraded.

\subsection{Contrast}

Contrast perturbation weakens the visual separation between foreground and background while preserving the original structural layout. We randomly select a subset of text, icons, or decorative elements and apply one or more transformations. Text contrast perturbation is similar to readability perturbation with contrast. We edit icons and other graphical components using a diffusion-based tool that adjusts their color distributions so that they blend more closely with surrounding areas. For each modification, we record the specific foreground–background pairs where contrast was intentionally reduced.

\subsection{Alignment}

Alignment perturbation introduces positional shifts that disrupt the structural consistency of the design. For each poster, we randomly select a small set of elements and shift them horizontally or vertically so that they break expected alignment patterns such as column structure or centered grouping. When a text block has an element, that is not the overall background, just below it, we move both together so that the perturbation affects alignment without introducing unintended overlap or contrast issues. These modifications do not map cleanly onto specific localised pairs but instead makes changes to the poster on a global level.

\subsection{Overlap}

Overlap perturbation introduces unintended occlusions between non-text elements while keeping all other aspects of the layout unchanged. We randomly select a small number of icons, shapes, or decorative components and position them so that they partially cover other elements while maintaining the overall alignment. This produced cases such as icon overlapping with icon or shape overlapping with another object, hence reducing visual separability. For each overlap introduced, we record the specific element pairs involved for explicit supervision.

\section{Prompting Setup for Model Sensitivity}
\label{sec:sensitivity_prompts}

To evaluate model sensitivity to PRISM perturbations (Section 4.3), 
we query each model under a consistent rubric-based prompting setup. The system prompt defines the scoring criteria for all five design principles and the overall score, each on a continuous scale between 0 and 1. The user prompt requests that the model return a JSON response without any explanation. For every composed or perturbed poster, the corresponding image is supplied directly to the model using its multimodal interface. For reproducibility, we include the exact prompts used for GPT-4o, GPT-4o-mini, and Qwen-2.5-VL in Figure 9.
We also include average scores and sensitivity ($\Delta$) from each model across design principles in Table 4.

\begin{figure}[t!]
     \centering
     \includegraphics[width=0.9\linewidth, trim={8.4cm 0cm 8.4cm 0cm},clip]{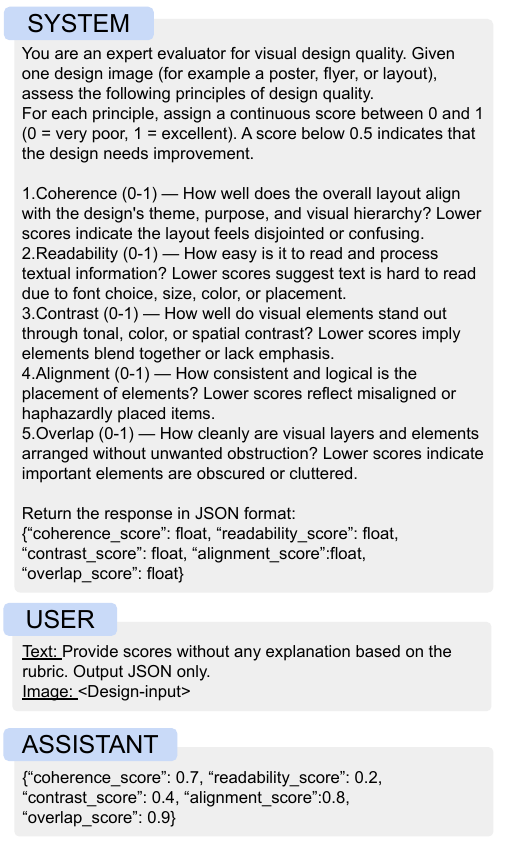}
     \caption{Prompts for evaluating model sensitivity to PRISM Perturbations. For results, see Fig 3. 
     }
     \label{fig:llm_prism}
 \end{figure}
\begin{table}[t!]
\centering
\small
\setlength{\tabcolsep}{4pt}
\renewcommand{\arraystretch}{0.9}
\resizebox{\linewidth}{!}{
\begin{tabular}{l l c c c c}
\toprule
 & \textbf{Backbone} & \textbf{Precision} $\uparrow$ 
& \textbf{Recall} $\uparrow$ & \textbf{F1-score} $\uparrow$ & \textbf{AUC} $\uparrow$ \\
\midrule
\multirow{4}{*}{\textbf{Frozen}} 
& ViT-B/16        & 0.528 & 0.522 & 0.524 & 0.554 \\
& DINOv2          & 0.539 & 0.531 & 0.535 & 0.563 \\
& OpenCLIP ViT-B/16   & 0.551 & 0.544 & 0.547 & 0.577 \\
& SigLIP-v2       & 0.556 & 0.548 & 0.552 & 0.584 \\
\midrule
\multirow{2}{*}{\textbf{Fine-tuned}} 
& OpenCLIP ViT-B/16   & 0.701 & 0.693 & 0.697 & 0.781 \\
& \cellcolor{cyan!10}\textbf{SigLIP-v2 (Ours)} & \cellcolor{cyan!10}\textbf{0.7391} & \cellcolor{cyan!10}\textbf{0.7382} & \cellcolor{cyan!10}\textbf{0.7386} & \cellcolor{cyan!10}\textbf{0.8126} \\
\bottomrule
\end{tabular}
}
\caption{\textbf{Backbone comparison for PRISM-scorer.}
Untrained backbones show limited design-sensitivity, while pretrained contrastive models 
(CLIP, SigLIP-v2) demonstrate strong performance. 
\textbf{SigLIP-v2 (Ours)} achieves the best overall scores across all metrics.
}
\label{tab:backbone_simple}
\end{table}

\begin{table*}[t]
\centering

\begin{subtable}{\textwidth}
\centering
\resizebox{\textwidth}{!}{
\begin{tabular}{@{}llccclccclccclccclccc@{}}
\toprule
\multirow{2}{*}{\textbf{Model}} &  & \multicolumn{3}{c}{\textbf{Coherence}} &  & \multicolumn{3}{c}{\textbf{Readability}} &  & \multicolumn{3}{c}{\textbf{Contrast}} &  & \multicolumn{3}{c}{\textbf{Alignment}} &  & \multicolumn{3}{c}{\textbf{Overlap}} \\ \cmidrule(lr){3-5} \cmidrule(lr){7-9} \cmidrule(lr){11-13} \cmidrule(lr){15-17} \cmidrule(l){19-21}  
&  & Org        & Perturb      & $\Delta$      &  & Org         & Perturb       & $\Delta$      &  & Org        & Perturb      & $\Delta$     &  & Org        & Perturb      & $\Delta$      &  & Org        & Perturb     & $\Delta$     \\ \midrule
Qwen-2.5-VL                     &  & 0.739      & 0.678        & 0.061      &  & 0.807       & 0.736         & 0.071      &  & 0.742      & 0.684        & 0.058     &  & 0.780      & 0.690        & 0.090      &  & 0.718      & 0.669       & 0.049     \\
GPT-4o-mini                     &  & 0.751      & 0.662        & 0.088      &  & 0.678       & 0.597         & 0.081      &  & 0.635      & 0.590        & 0.045     &  & 0.726      & 0.622        & 0.104      &  & 0.782      & 0.691       & 0.091     \\
GPT-4o                          &  & 0.824      & 0.683        & 0.141      &  & 0.693       & 0.533         & 0.160      &  & 0.759      & 0.671        & 0.087     &  & 0.827      & 0.689        & 0.138      &  & 0.787      & 0.619       & 0.168     \\
\rowcolor{cyan!10} \textbf{PRISM-Scorer}                    &  & 0.658      & 0.377        & \textbf{0.281}      &  & 0.633       & 0.602         & \textbf{0.031}      &  & 0.712      & 0.638        & \textbf{0.074}     &  & 0.743      & 0.621        & \textbf{0.122}      &  & 0.655      & 0.589       & \textbf{0.066 }    \\ \bottomrule
\end{tabular}
}
\caption{\textbf{Coherence Results.}}
\end{subtable}

\begin{subtable}{\textwidth}
\resizebox{\textwidth}{!}{
\begin{tabular}{@{}llccclccclccclccclccc@{}}
\toprule
\multirow{2}{*}{\textbf{Model}} &           & \multicolumn{3}{c}{\textbf{Coherence}}                    &           & \multicolumn{3}{c}{\textbf{Readability}}                  &           & \multicolumn{3}{c}{\textbf{Contrast}}                     &           & \multicolumn{3}{c}{\textbf{Alignment}}                    &           & \multicolumn{3}{c}{\textbf{Overlap}}                      \\ \cmidrule(lr){3-5} \cmidrule(lr){7-9} \cmidrule(lr){11-13} \cmidrule(lr){15-17} \cmidrule(l){19-21} 
&           & Org            & Perturb        &$\Delta$          &           & Org            & Perturb        &$\Delta$          &           & Org            & Perturb        &$\Delta$          &           & Org            & Perturb        &$\Delta$          &           & Org            & Perturb        &$\Delta$          \\ \midrule
Qwen-2.5-VL            &           & 0.744          & 0.690          & 0.054          &           & 0.830          & 0.725          & 0.105          &           & 0.751          & 0.689          & 0.062          &           & 0.795          & 0.703          & 0.092          &           & 0.736          & 0.681          & 0.055          \\
GPT-4o-mini            &           & 0.751          & 0.680          & 0.071          &           & 0.675          & 0.575          & 0.100          &           & 0.633          & 0.580          & 0.053          &           & 0.734          & 0.580          & 0.154          &           & 0.784          & 0.720          & 0.064          \\
GPT-4o                 &           & 0.822          & 0.704          & 0.118          &           & 0.709          & 0.459          & 0.250          &           & 0.784          & 0.626          & 0.158          &           & 0.826          & 0.705          & 0.121          &           & 0.795          & 0.604          & 0.190          \\
\rowcolor{cyan!10} \textbf{PRISM-Scorer}  & & 0.599 & 0.543 & \textbf{0.056} & & 0.664 & 0.340 & \textbf{0.324} & & 0.712 & 0.578 & \textbf{0.134} & & 0.643 & 0.541 & \textbf{0.102} & & 0.725 & 0.645 & \textbf{0.080} \\ \bottomrule
\end{tabular}
}
\centering\caption{\textbf{Readability Results.}}
\end{subtable}

\begin{subtable}{\textwidth}
\resizebox{\textwidth}{!}{
\begin{tabular}{@{}llrrrlrrrlrrrlrrrlrrr@{}}
\toprule
\multirow{2}{*}{\textbf{Model}} &  & \multicolumn{3}{c}{\textbf{Coherence}}                                            &  & \multicolumn{3}{c}{\textbf{Readability}}                                          &  & \multicolumn{3}{c}{\textbf{Contrast}}                                             &  & \multicolumn{3}{c}{\textbf{Alignment}}                                            &  & \multicolumn{3}{c}{\textbf{Overlap}}                                              \\ \cmidrule(lr){3-5} \cmidrule(lr){7-9} \cmidrule(lr){11-13} \cmidrule(lr){15-17} \cmidrule(l){19-21} 
&  & \multicolumn{1}{c}{Org} & \multicolumn{1}{c}{Perturb} & \multicolumn{1}{c}{$\Delta$} &  & \multicolumn{1}{c}{Org} & \multicolumn{1}{c}{Perturb} & \multicolumn{1}{c}{$\Delta$} &  & \multicolumn{1}{c}{Org} & \multicolumn{1}{c}{Perturb} & \multicolumn{1}{c}{$\Delta$} &  & \multicolumn{1}{c}{Org} & \multicolumn{1}{c}{Perturb} & \multicolumn{1}{c}{$\Delta$} &  & \multicolumn{1}{c}{Org} & \multicolumn{1}{c}{Perturb} & \multicolumn{1}{c}{$\Delta$} \\ \midrule
Qwen-2.5-VL                     &  & 0.751                   & 0.691                       & 0.06                      &  & 0.810                   & 0.730                       & 0.08                      &  & 0.720                   & 0.610                       & 0.11                      &  & 0.720                   & 0.635                       & 0.085                     &  & 0.720                   & 0.660                       & 0.06                      \\
GPT-4o-mini                     &  & 0.752                   & 0.677                       & 0.075                     &  & 0.630                   & 0.535                       & 0.095                     &  & 0.610                   & 0.460                       & 0.15                      &  & 0.735                   & 0.625                       & 0.11                      &  & 0.810                   & 0.735                       & 0.075                     \\
GPT-4o                          &  & 0.810                   & 0.690                       & 0.12                      &  & 0.720                   & 0.520                       & 0.2                       &  & 0.790                   & 0.540                       & 0.25                      &  & 0.790                   & 0.660                       & 0.13                      &  & 0.750                   & 0.590                       & 0.16                      \\
\rowcolor{cyan!10} \textbf{PRISM-Scorer}           &  & 0.620                   & 0.550                       & \textbf{0.07}             &  & 0.610                   & 0.415                       & \textbf{0.195}            &  & 0.710                   & 0.400                       & \textbf{0.31}             &  & 0.670                   & 0.565                       & \textbf{0.105}            &  & 0.710                   & 0.625                       & \textbf{0.085}            \\ \bottomrule
\end{tabular}
}
\caption{\textbf{Contrast Results.}}
\end{subtable}

\begin{subtable}{\textwidth}
\resizebox{\textwidth}{!}{
\begin{tabular}{@{}llrrrlrrrlrrrlrrrlrrr@{}}
\toprule
\multirow{2}{*}{\textbf{Model}} &  & \multicolumn{3}{c}{\textbf{Coherence}}                                            &  & \multicolumn{3}{c}{\textbf{Readability}}                                          &  & \multicolumn{3}{c}{\textbf{Contrast}}                                             &  & \multicolumn{3}{c}{\textbf{Alignment}}                                            &  & \multicolumn{3}{c}{\textbf{Overlap}}                                              \\ \cmidrule(lr){3-5} \cmidrule(lr){7-9} \cmidrule(lr){11-13} \cmidrule(lr){15-17} \cmidrule(l){19-21} 
&  & \multicolumn{1}{c}{Org} & \multicolumn{1}{c}{Perturb} & \multicolumn{1}{c}{$\Delta$} &  & \multicolumn{1}{c}{Org} & \multicolumn{1}{c}{Perturb} & \multicolumn{1}{c}{$\Delta$} &  & \multicolumn{1}{c}{Org} & \multicolumn{1}{c}{Perturb} & \multicolumn{1}{c}{$\Delta$} &  & \multicolumn{1}{c}{Org} & \multicolumn{1}{c}{Perturb} & \multicolumn{1}{c}{$\Delta$} &  & \multicolumn{1}{c}{Org} & \multicolumn{1}{c}{Perturb} & \multicolumn{1}{c}{$\Delta$} \\ \midrule
Qwen-2.5-VL                     &  & 0.748                   & 0.690                       & 0.058                     &  & 0.804                   & 0.732                       & 0.072                     &  & 0.734                   & 0.667                       & 0.067                     &  & 0.751                   & 0.611                       & 0.14                      &  & 0.717                   & 0.652                       & 0.065                     \\
GPT-4o-mini                     &  & 0.720                   & 0.638                       & 0.082                     &  & 0.652                   & 0.562                       & 0.09                      &  & 0.641                   & 0.561                       & 0.08                      &  & 0.742                   & 0.542                       & 0.2                       &  & 0.827                   & 0.737                       & 0.09                      \\
GPT-4o                          &  & 0.815                   & 0.690                       & 0.125                     &  & 0.731                   & 0.566                       & 0.165                     &  & 0.781                   & 0.661                       & 0.12                      &  & 0.803                   & 0.593                       & 0.21                      &  & 0.731                   & 0.576                       & 0.155                     \\
\rowcolor{cyan!10} \textbf{PRISM-Scorer}           &  & 0.652                   & 0.582                       & \textbf{0.07}             &  & 0.640                   & 0.490                       & \textbf{0.15}             &  & 0.721                   & 0.611                       & \textbf{0.11}             &  & 0.692                   & 0.382                       & \textbf{0.31}             &  & 0.728                   & 0.558                       & \textbf{0.17}             \\ \bottomrule
\end{tabular}
}
\caption{\textbf{Alignment Results.}}
\end{subtable}

\begin{subtable}{\textwidth}
\resizebox{\textwidth}{!}{
\begin{tabular}{@{}llrrrlrrrlrrrlrrrlrrr@{}}
\toprule
\multirow{2}{*}{\textbf{Model}} &  & \multicolumn{3}{c}{\textbf{Coherence}}                                            &  & \multicolumn{3}{c}{\textbf{Readability}}                                          &  & \multicolumn{3}{c}{\textbf{Contrast}}                                             &  & \multicolumn{3}{c}{\textbf{Alignment}}                                            &  & \multicolumn{3}{c}{\textbf{Overlap}}                                              \\ \cmidrule(lr){3-5} \cmidrule(lr){7-9} \cmidrule(lr){11-13} \cmidrule(lr){15-17} \cmidrule(l){19-21} 
&  & \multicolumn{1}{c}{Org} & \multicolumn{1}{c}{Perturb} & \multicolumn{1}{c}{$\Delta$} &  & \multicolumn{1}{c}{Org} & \multicolumn{1}{c}{Perturb} & \multicolumn{1}{c}{$\Delta$} &  & \multicolumn{1}{c}{Org} & \multicolumn{1}{c}{Perturb} & \multicolumn{1}{c}{$\Delta$} &  & \multicolumn{1}{c}{Org} & \multicolumn{1}{c}{Perturb} & \multicolumn{1}{c}{$\Delta$} &  & \multicolumn{1}{c}{Org} & \multicolumn{1}{c}{Perturb} & \multicolumn{1}{c}{$\Delta$} \\ \midrule
Qwen-2.5-VL                     &  & 0.732                   & 0.672                       & 0.06                      &  & 0.821                   & 0.746                       & 0.075                     &  & 0.730                   & 0.660                       & 0.07                      &  & 0.756                   & 0.671                       & 0.085                     &  & 0.750                   & 0.610                       & 0.14                      \\
GPT-4o-mini                     &  & 0.762                   & 0.677                       & 0.085                     &  & 0.625                   & 0.525                       & 0.1                       &  & 0.658                   & 0.563                       & 0.095                     &  & 0.752                   & 0.637                       & 0.115                     &  & 0.783                   & 0.598                       & 0.185                     \\
GPT-4o                          &  & 0.809                   & 0.679                       & 0.13                      &  & 0.713                   & 0.533                       & 0.18                      &  & 0.774                   & 0.634                       & 0.14                      &  & 0.832                   & 0.682                       & 0.15                      &  & 0.784                   & 0.544                       & 0.24                      \\
\rowcolor{cyan!10} \textbf{PRISM-Scorer}           &  & 0.630                   & 0.555                       & \textbf{0.075}            &  & 0.656                   & 0.466                       & \textbf{0.19}             &  & 0.741                   & 0.611                       & \textbf{0.13}             &  & 0.643                   & 0.533                       & \textbf{0.11}             &  & 0.732                   & 0.412                       & \textbf{0.32}             \\ \bottomrule
\end{tabular}
}
\caption{\textbf{Overlap Results.}}
\end{subtable}
\caption{\textbf{Average Model Sensitivity Scores.} The table shows the scores from different models on original vs perturbed layouts across design principles, $\Delta$ denotes the difference between the averages showing sensitivity. While GPT-4o shows sensitivity, PRISM-Scorer is able to disentangle the principles better than any other model.}
\label{tab:delta_tables}
\end{table*}

\section{Backbone Ablation}
\label{sec:backbone}
Table 3 compares several visual backbones. This ablation evaluates whether architectural choice or multimodal pretraining affects a model’s ability to detect principle-specific design degradations. In the frozen setting, each backbone is initialized with its standard pretrained weights: ViT-B/16 \cite{dosovitskiy2021imageworth16x16words} trained on ImageNet \cite{russakovsky2015imagenetlargescalevisual}, SigLIP-v2 \cite{siglipv2} trained on WebLI \cite{webli}, and the self-supervised DINOv2 model \cite{oquab2024dinov2learningrobustvisual}, but we freeze the entire backbone and train only an identical lightweight two-layer binary classifier on top for all models, ensuring that differences arise solely from the backbone representations. We then report averaged PRISM-scorer performance across all five principles. These frozen models provide limited but non-trivial sensitivity to design perturbations, with vision-only backbones (ViT-B/16, DINOv2) performing the weakest and frozen vision–language architectures (OpenCLIP \cite{ilharco_gabriel_2021_5143773}, SigLIP-v2) offering slightly stronger baselines due to their multimodal structure. In the fine-tuned setting, we update all backbone weights using the principle-aware fine-tuning procedure described in 
Section 3.2.1, leveraging paired image–caption data from Crello \cite{canvasvae} and CreatiDesign \cite{zhang2025creatidesignunifiedmulticonditionaldiffusion}. This multimodal contrastive supervision leads to substantial improvements: fine-tuned OpenCLIP ViT-B/16 significantly outperforms all frozen variants, and SigLIP-v2 (Ours) achieves the strongest performance across precision, recall, F1, and AUC. These findings underscore the importance of principle-aware multimodal fine-tuning and motivate our use of SigLIP-v2 as the backbone for PRISM-scorer. 

We also compare the simple linear head on top of this design-aware architecture with other variants. Averaged across principles, the test F1 scores are: linear head (\SI{73.6}{\percent}), two-layer MLP (\SI{71.4}{\percent}), and multi-head attention (\SI{72.5}{\percent}). These results show that a simpler linear layer is sufficient to achieve the required performance gains. 

We compare a unified scorer–localizer model, evaluated by training Qwen-2.5-VL end-to-end. The unified model achieves an average F1 score of \SI{65}{\percent} as a scorer and mean IoU scores of \SI{60}{\percent} across principles. The unified model performs worse than our expert PRISM scorer and localizer models. This explains the need for experts instead of one unified model trained for multiple tasks.

\begin{figure*}[t]
     \centering
     \includegraphics[width=\linewidth, trim={0cm 0cm 1cm 0cm},clip]{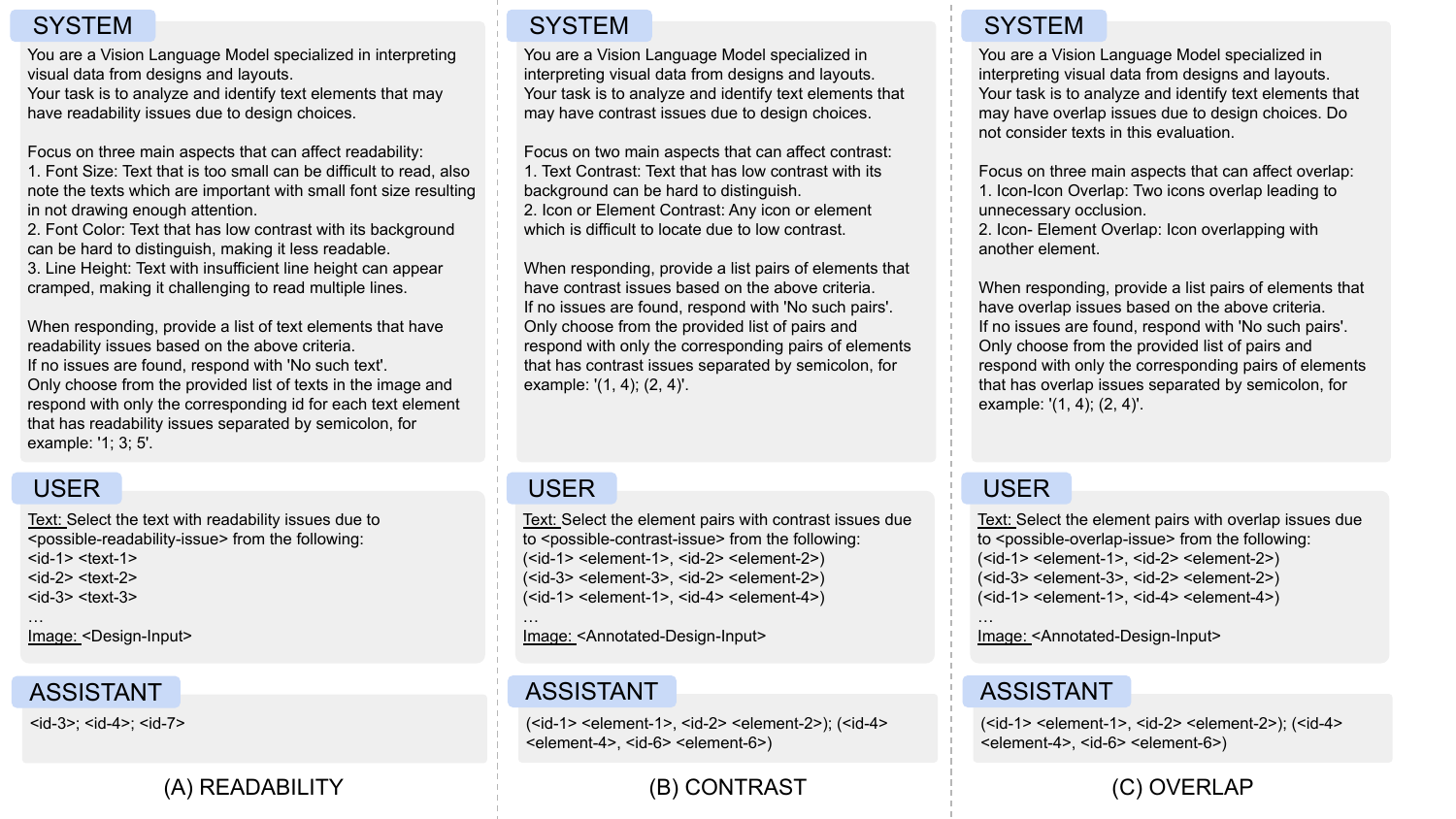}
     \caption{\textbf{Prompts for Localising Error Detection.} The figure shows the prompts used for instruct-tuning and evaluating models for local principles (Readability (A), Contrast (B), Overlap (C)). Results comparing different models using this prompt can be seen in Table 2.
     }
     \label{fig:localiser_prompts}
 \end{figure*}
\section{Localised Error Detection Details}
\label{sec:localised_details}
We get the data required for instruction-tuning as described in Section~\ref{sec:perturbation_details}. To train the model on localised supervision from readability, contrast and overlap perturbations, we use the prompts as shown in Figure~\ref{fig:localiser_prompts}. For readability, the corresponding image input is simply the design, whereas for contrast and overlap, we annotate the design using the metadata to include IDs for each non-textual element. For readability, the model is asked to choose from a list of texts present in the design, while for contrast and overlap, we provide pairs of elements to choose from. The list only contains pairs which have some overlapping components in order to contain the combinations to a reasonable number. We use the same prompt for evaluating and comparing different models on the held-out set. We use the training and evaluating methods as mentioned in 
Section 4.4 and 5.3, respectively, for the results refer to Figure 2. 
\begin{figure*}[t]
     \centering
     \includegraphics[width=0.95\linewidth, trim={0cm 3cm 0cm 2.5cm},clip]{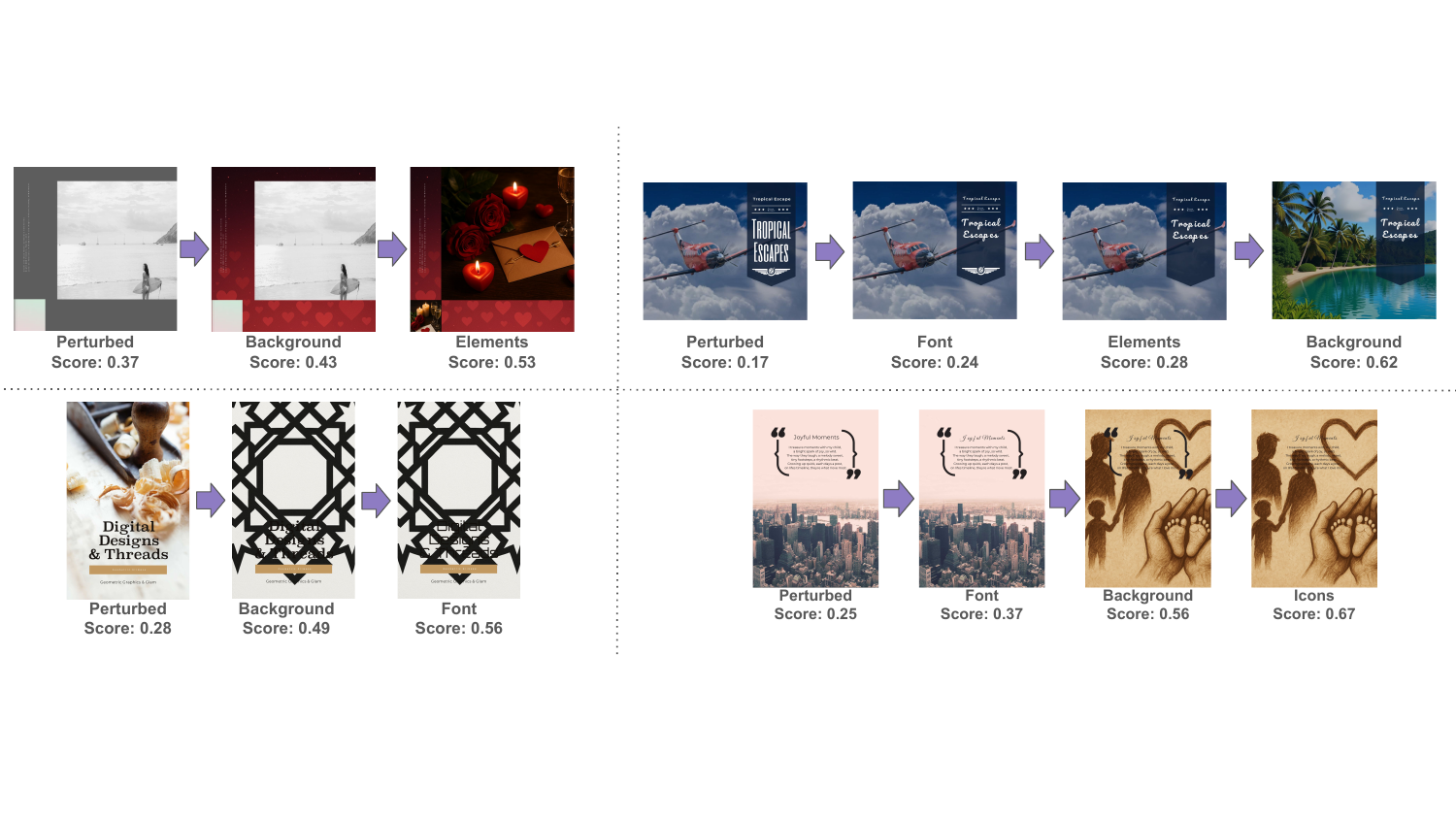}
     \caption{\textbf{Coherence-based editing examples.} The grid shows perturbed posters and the editing path (See Section 5.4).
     Since the pipeline focuses only on coherence, edits target the background, elements, and font for thematic consistency, while other principles remain unchanged. Scores reflect the scorer’s coherence predictions at each step.}
     \label{fig:more_edited}
 \end{figure*}

\section{Scorer Out-of-domain generalization} 
\label{sec:OOD}
As observed in Figure 7, the PRISM scorer is able to disentangle design principles. We evaluate the coherence scorer with all other perturbations that it has not seen during training, giving us insights into the scorer's generalization performance. While the scorer can generalize to all types of PRISM perturbations,  we test the potential of the scorer on a different perturbation style from the GDE dataset \cite{gde2}. This dataset only contains overlap and alignment perturbations. We evaluate our scorers using human annotations from the GDE dataset as ground truth. The PRISM scorer for overlap and alignment achieves F1 scores of \SI{72}{\percent} and \SI{71}{\percent}, respectively, showing generalization to other types of perturbations as well. 

\section{Additional Examples from Editing Pipeline}
\label{sec:more_examples}
As shown in Figure~\ref{fig:more_edited}, we present additional examples from our demonstrative editing pipeline, which operates exclusively using the coherence module. Candidate edits are proposed by the coherence localizer and ranked at each step by the coherence scorer, resulting in refinements that focus solely on restoring thematic and stylistic consistency (See Section 5.4).
These examples only look at coherence-driven changes, while keeping the other principles untouched. A natural next step is to extend the editing pipeline to all design principles and explore ways to combine these principle-specific refinements. A unified system that can coordinate edits across principles would enable more comprehensive and well-rounded layout improvement.

\end{document}